\documentclass{article}

\usepackage[preprint, nonatbib]{neurips_2022}

\usepackage[utf8]{inputenc} 
\usepackage[T1]{fontenc}    
\usepackage{hyperref}       
\usepackage{url}            
\usepackage{booktabs}       
\usepackage{amsfonts}       
\usepackage{nicefrac}       
\usepackage{microtype}      
\usepackage[table,xcdraw]{xcolor}
\usepackage{graphicx}       
\usepackage{tcolorbox}
\usepackage{multirow}
\usepackage{rotating}
\usepackage{adjustbox}
\usepackage{pifont}
\usepackage{longtable}
\usepackage{lscape}
\usepackage{docmute}
\usepackage{soul}
\usepackage{tikz}
\usepackage{caption}
\usepackage{subcaption}
\usetikzlibrary{fit, positioning}
\usepackage[
backend=bibtex,
maxbibnames=20,
sorting=none,
date=year,
doi=false,
isbn=false,
url=false
]{biblatex}
\newcolumntype{C}[1]{>{\centering\arraybackslash}p{#1}}

\newcommand{\hlc}[2][red!40!white]{{%
    \colorlet{foo}{#1}%
    \sethlcolor{foo}\hl{#2}}%
}

\title{ChatGPT Makes Medicine Easy to Swallow:\\An Exploratory Case Study on Simplified\\Radiology Reports}

\author{%
\textbf{Katharina Jeblick}$^{1,2}$ \quad
\textbf{Balthasar Schachtner}$^{1}$ \quad
\textbf{Jakob Dexl}$^{1,3}$ \quad 
\textbf{Andreas Mittermeier}$^{1}$  \quad \\
\textbf{Anna Theresa Stüber}$^{1,4}$ \quad 
\textbf{Johanna Topalis}$^{1}$  \quad 
\textbf{Tobias Weber}$^{1,3,4}$ \quad 
\textbf{Philipp Wesp}$^{1}$ \quad \\
\textbf{Bastian Sabel}$^{1}$ \quad 
\textbf{Jens Ricke}$^{1}$ \quad
\textbf{Michael Ingrisch}$^{1,3}$ \quad
\\
$^1$Department of Radiology, University Hospital, LMU Munich \\ 
\quad $^2$Comprehensive Pneumology Center (CPC-M),  Member of the \\German Center for Lung Research (DZL), Munich \\ 
\quad $^3$Munich Center for Machine Learning (MCML) \\
\quad $^4$Department of Statistics, LMU Munich\\
\texttt{\{katharina.jeblick, balthasar.schachtner\}@med.uni-muenchen.de}}

\begin{document}

\maketitle


\begin{abstract}
The release of ChatGPT, a language model capable of generating text that appears human-like and authentic, has gained significant attention beyond the research community.
We expect that the convincing performance of ChatGPT incentivizes users to apply it to a variety of downstream tasks, including prompting the model to simplify their own medical reports.
To investigate this phenomenon, we conducted an exploratory case study.
In a questionnaire, we asked 15 radiologists to assess the quality of radiology reports simplified by ChatGPT.
Most radiologists agreed that the simplified reports were factually correct, complete, and not potentially harmful to the patient.
Nevertheless, instances of incorrect statements, missed key medical findings, and potentially harmful passages were reported.
While further studies are needed, the initial insights of this study indicate a great potential in using large language models like ChatGPT to improve patient-centered care in radiology and other medical domains.
\begingroup
\hypersetup{hidelinks}
{\let\thefootnote\relax\footnote{{All authors contributed equally.}}}
{\let\thefootnote\relax\footnote{{}}}
{\let\thefootnote\relax\footnote{{The title was generated with ChatGPT.}}}
\endgroup
\setcounter{footnote}{0}
\end{abstract}


\section{Introduction}
"ChatGPT, am I dying? What does this medical report mean? Can you explain it to me like I'm five?"
With the latest release of OpenAI's Large Language Model (LLM) ChatGPT \cite{chatgpt}, algorithmic language modeling has reached a new milestone in generating human-like responses to user text inputs.
Just five days after its release, ChatGPT had already attracted over a million users and gained a significant amount of media attention \cite{nyt, wapo, bbc, guardian}.
Given the level of popularity and widespread access to the public, the question as to how people will use such models arises and which opportunities and challenges are associated with them.
In our experience, the text output of ChatGPT was astonishingly convincing for a variety of tasks, such that we expect disruptive change across numerous domains and industries in a very short time. 

Among a myriad of potential downstream applications, LLMs are able to simplify complex text \cite{agarwal2022explain} and subsequently make it more accessible to a broader audience.
There is a huge need for simplification in domains where expert knowledge is necessary to put text into context and interpret it.
For example, legal or medical documents are written by trained experts in highly specialized language, but often have immediate consequences for clients or patients, respectively, who are usually non-experts. 
Simplifying these documents promises to enable individuals to actively and autonomously oversee them.
One domain where text simplification might be particularly useful is radiology.
Radiological findings are typically communicated in a free-text report in specialized medical jargon, targeting a clinician or doctor as the recipient.
For readers without a medical background, these reports are often inaccessible and potentially misleading.
For example, a study \cite{martin-carreras_readability_2019} showed, that only 4\% of all analyzed radiology reports were readable by the average US adult. They concluded to use simpler and more structured language to improve patient-centered care.

Ideally, radiological findings are communicated in a timely, personalized conversation between doctor and patient.
However, in resource-constrained health systems, this conversation is often delayed.
Therefore, patients might try to make their radiology reports accessible upfront.
For instance, patients might research isolated medical terms on the internet without being able to place them in the correct medical context, potentially leading to misinterpretations and confusion. In this context, \cite{oh2016porter} developed an online system that augments radiology reports with lay-language definitions.
Alternatively, patients might use free simplification services\footnote{https://washabich.de (visited on 12/28/2022)}, which often lack popularity, accessibility, or scalability.
We expect, that in the very near future, publicly accessible LLMs such as ChatGPT will increasingly be applied by interested patients to simplify their own medical reports filling the gap. 

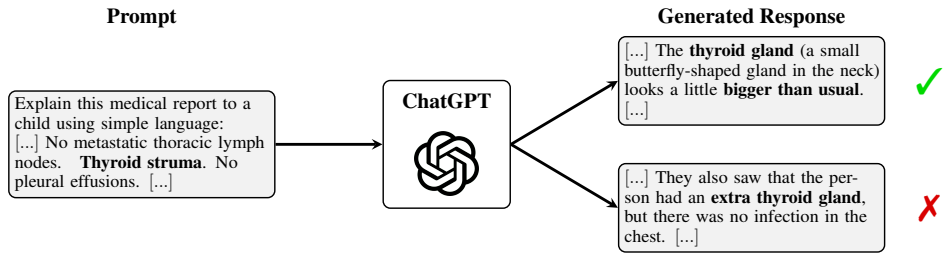
\begin{figure}
    \centering
    \resizebox{0.9\textwidth}{!}{%
    	\begin{tikzpicture}
		\node[] at (0, 0) (a) {\large \textbf{ChatGPT}};
		\node[below of = a, yshift=-.25cm] (b) {\includegraphics[width=1.25cm]{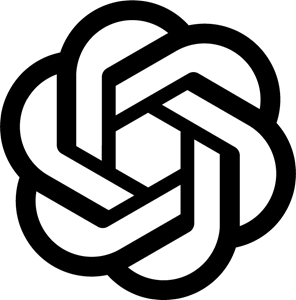}};
		\node[draw, fit=(a) (b), thick, rounded corners, minimum width=2.5cm, minimum height=2.5cm](c) {};
		
		\node[text width=5cm, draw, thick, fill=gray!10!white, left of= c, xshift=-5cm, rounded corners] (d) {
			Explain this medical report to a child using simple language:
			
			$\lbrack ... \rbrack$
			No metastatic thoracic lymph nodes. \textbf{ Thyroid struma}. No pleural effusions.
			$\lbrack ... \rbrack$
		};
	
		\node[left of= c, xshift=-5cm, yshift=2.5cm] {\large \textbf{Prompt}};
		
		\node[right of= c, xshift=5cm, yshift=2.5cm] {\large \textbf{Generated Response}};

		\node[text width=5cm, draw, thick, fill=gray!10!white, right of= c, xshift=5cm, yshift=-1.25cm, rounded corners] (e) {
		$\lbrack ... \rbrack$
        They also saw that the person had an \textbf{extra thyroid gland}, but
there was no infection in the chest.
		$\lbrack ... \rbrack$
	};
		
		\node[text width=5cm, draw, thick, fill=gray!10!white, right of= c, xshift=5cm, yshift=1.25cm, rounded corners] (f) {
		$\lbrack ... \rbrack$
		          The \textbf{thyroid gland} (a small butterfly-shaped gland in the neck) looks a little \textbf{bigger than usual}.
		$\lbrack ... \rbrack$
	};

		\node[left of= f, xshift=4.5cm] (g) {{\huge \color{green!85!black}\ding{51}}};
		\node[left of= e, xshift=4.5cm] (g) {{\huge \color{red!85!black}\ding{55}}};

	\draw [-stealth, ultra thick](c.east) -- (f.west);
	\draw [-stealth, ultra thick](c.east) -- (e.west);
	\draw [stealth-, ultra thick](c.west) -- (d.east);
	\end{tikzpicture}
	}
    \caption{Prompting ChatGPT to simplify a radiology report of an oncological CT. While producing plausible-sounding responses, the content does not need to be true.}
    \label{fig:chatgpt-intro}
\end{figure}

However, ChatGPT was not explicitly trained for medical text simplification and is not intended to be used for this critical task. While it can generate plausible-sounding text, the content does not need to be true, as shown in Figure~\ref{fig:chatgpt-intro}. 
This raises the question of whether LLMs such as ChatGPT are able to simplify radiology reports such that the output is factually correct, complete, and not potentially harmful to the patient.
Patients, i.e., potential LLM users, are not able to answer this question. The quality of these simplified reports needs to be thoroughly assessed by expert radiologists.

In this work, we conducted an exploratory case study to investigate the phenomenon that emerging LLMs such as ChatGPT may be used or misused by patients to simplify their radiology reports. 
This unintended use may raise chances and challenges for patient-centered care. 
Therefore, we asked 15 radiologists to rate the quality of three radiology reports simplified by ChatGPT regarding their \textit{factual correctness}, \textit{completeness}, and \textit{potential harm} using a questionnaire.
Descriptive statistics and inductive categories were used to evaluate the questionnaire. 
Based on our findings, we elaborated opportunities and challenges of using ChatGPT-like LLMs for simplifying radiology reports.

\section{Background}
In this section, we introduce LLMs, as well as the technical specifics of ChatGPT. 
We shortly discuss techniques for summarization and simplification of radiology reports and mention known limitations of LMMs. This will serve as a basis for discussing chances and potential challenges for a scenario where these models are used to simplify radiology reports by patients.

\subsection{Large Language Models in Natural Language Processing}
Since introducing the concept of attention in deep learning models \cite{vaswani2017attention}, the transformer is an established architecture with dominance in nearly all natural language processing (NLP) benchmarks, including question answering, translation, and text classification \cite{chowdhery2022palm}.
While the base-model architecture remained relatively unchanged throughout the years, significant progress was made by scaling the number of layers and internal dimensions resulting in so-called Large Language Models (LLMs) with billions of parameters, which lead to increased model capacity and abilities.
Popular architectures grew in parameter size starting, e.g., from BERT (345M; \cite{devlin2018bert}) over MegatronLM (8.3B; \cite{shoeybi2019megatron}) and T-5 (11B; \cite{raffel2020exploring}) up to BLOOM (176B; \cite{scao2022bloom}) and PaLM (540B; \cite{chowdhery2022palm}).
A significant success factor for fitting LLMs is an enormous training dataset, e.g., the Pile \cite{gao2020pile}, which contains documents from Arxiv, PubMed, Stack Exchange, Wikipedia, as well as a subset of Common Crawl\footnote{http://commoncrawl.org/}, and GitHub, among others.
For these kinds of LLMs, \cite{bommasani2021opportunities} introduced the terminology of \textit{foundation models}, which defines training on a very large data basis and the ability to adapt to a variety of downstream tasks.  

\subsection{ChatGPT}
ChatGPT is an LLM  developed by OpenAI that was first released on November 30th, 2022.
The user can directly prompt the model via an API in a conversational way, e.g., allowing for follow-up questions or admission of mistakes \cite{chatgpt}.
The backbone of ChatGPT is based on the generative pre-trained transformer series (GPT; \cite{radford2018improving, radford2019language, brown2020language}).
Despite the success and capacity of the third GPT iteration (GPT-3) \cite{brown2020language} with 175B parameters, the challenge of engineering text prompts for achieving the desired generative output remained.
This is due to the autoregressive training procedure, which tasks the model to predict a token based on the previous text and thus is optimized for text completion and not dialogues.
To improve the dialogue capabilities of the model as well as to reduce bias and general toxicity, \cite{ouyang2022training} developed InstructGPT, a fine-tuned GPT-3, that leverages a novel training procedure in order to follow user instructions more persistently.
The basic idea for InstructGPT is reinforcement learning from human feedback (RLHF, \cite{christiano2017deep}), which is composed of three phases:
First, human labelers manually create responses for randomly sampled prompts from a prompt database.
This handcrafted data served to fine-tune GPT-3 in a supervised fashion (SFT model), i.e., through training on the manually created output for the respective prompt.
In the second step, SFT is tasked to create multiple outputs for a given prompt, which are subsequently ranked by a human labeler from best to worst.
This new ranking dataset serves to train a reward model, which assigns a scalar reward as a measure of quality for the given SFT outputs.
In the final step, concepts from reinforcement learning and proximal policy optimization \cite{schulman2017proximal} were applied to fine-tune SFT using a previously trained reward model as an approximate reward function.
This allows evaluation of the quality of the online-generated outputs using the reward model.

The setups of training InstructGPT and ChatGPT are nearly identical with minor differences concerning the data collection \cite{chatgpt}.
The major difference lies in the backbone itself. InstructGPT is based on GPT-3.  ChatGPT uses GPT-3.5,  a newer, iterated version of the original GPT-3. 
Additionally, OpenAI embedded various non-disclosed safety mechanisms into ChatGPT to promote a secure and non-toxic environment.

\subsection{Simplification and Summarization of Radiology Reports}
Despite being closely related, simplification and summarization express two different concepts.
Text summarization describes the task of creating a short version of a text including all important aspects.
A supervised recurrent model for this domain was proposed by \cite{zhang2018learning} with increased performance in comparison to previous non-neural approaches.
\cite{cai2021chestxraybert} noted a lack of medical terminology in BERT and therefore proposed fine-tuning the model on chest x-ray reports before summarizing impression sections.
A domain adaption was done by \cite{liang2022fine} for a German version of BERT with the goal of summarizing German chest x-ray reports.
Further approaches for abstractive summarization employed, e.g., a pointer-generator network \cite{macavaney2019ontology} or a reinforcement-learning inspired procedure \cite{gigioli2018domain} to solve this task.
For extractive summarization of medical data, BioBERT was fine-tuned by \cite{du2020biomedical} and a sentence-ranking framework with random forests in the backbone was proposed by \cite{lee2020cerc}.
Overall, \cite{chaves2022automatic} argued that the continuously growing availability of biomedical text data has led to increased attention to the research field of text summarization, where meanwhile the majority of methods focus on machine learning.

In contrast to summarization, simplifying a text does not necessarily imply shortening, but describes a transformation to make it more readable and understandable \cite{shardlow2014survey, al2021automated}. For instance, 
\cite{kvist2013professional} identified a considerate proportion of recurring standard phrases in their radiology report dataset and argued that these could be easily and consistently replaced with simpler terminology, while more varied descriptions of unique pathological findings were less easy to simplify automatically.
A similar procedure was applied in \cite{abrahamsson2014medical}, where human evaluators report a simplification of the medical reports after the substitution.
The replacement method was also adapted and extended by a semantic network in \cite{ramadier2017radiological}.
The open-access consumer health vocabulary (CHV) was used by \cite{qenam2017text} for simplifying text. They came to the conclusion that even some CHV terms may not be appropriate for laypersons.

In addition to this technical work on how to generate simplified reports, studies were conducted to investigate how well patients can understand conventional radiology reports, and if they prefer simplified reports.
Using readability tests, \cite{yi_readability_2019} showed that the required reading level for their set of lumbar spine reports exceeded 12$^{th}$ grade. They argued that the patients' understanding of complex texts should be considered since patients increasingly read reports themselves.
As mentioned in the introduction, \cite{martin-carreras_readability_2019} concluded that only 4\% of more than 100,000 consecutive radiology reports were at a reading level below the reading level of an average US adult.
This finding was further investigated in a randomized controlled trial \cite{barrett_patient-centered_2021}, where the majority of patients showed a preference for patient-friendly radiology reports over traditional radiology reports. 

There is a clear need for patients to be able to better understand radiology reports \cite{martin-carreras_readability_2019}.
While machine learning has made significant progress in the area of radiology report summarization, to the best of our knowledge, automated report simplification has not yet received adequate attention within the NLP community.
ChatGPT and other foundation models present promising opportunities for simplifying radiology reports, but it is important to consider their limitations, particularly when used in the medical field.

\subsection{Known Limitations of LLMs}
The GPT-3 publication \cite{brown2020language} discussed the potential for misuse since the generated outputs become close to indistinguishable from human language.
Furthermore, they highlighted model-inherent issues with biases and fairness.
Access to GPT is only granted over an API and not to the original model.
With an effort to democratize artificial intelligence, EleutherAI published GPT-J as an open-source alternative for GPT-3 \cite{gpt-j}.
\cite{bommasani2021opportunities} also elaborated on intrinsic biases, which are inherent to foundation models. In addition, they brought up the concept of extrinsic harms that occur by adaption to a downstream task.
Moreover, the aspect of centralization of power was mentioned as a problem, where only a specific part of society is able to access tools like an LLM.
\cite{bender2021dangers} critically discussed the aspect of training data being one source of harmful and biased behavior of LLMs and depicted the financial as well as environmental impacts of large model training.
Additionally, \cite{carlini2021extracting} were able to extract potentially sensitive training data from LLMs, indicating that larger models tend to be more vulnerable to these attacks due to increased memorization.
Another important issue is that LLMs can produce a text which sounds plausible to the reader but is incorrect or nonsensical.
In a benchmark to measure truthful answers, larger models tended to score lower than their smaller versions with fewer parameters \cite{lin2021truthfulqa}.
Recently, Meta launched the Galactica LLM \cite{taylor2022galactica} to support scientific writing.
However, the API was taken offline three days after its release due to a backlash: The model produced seemingly correct-sounding but factually incorrect articles.

When using ChatGPT to simplify radiology reports, we expect the output to sound plausible, while its factual correctness and completeness are not guaranteed. In the next chapter, we present our approach for assessing the quality of ChatGPT-generated simplified reports.

\section{Methods}

\begin{figure}[t]
    \centering
    \includegraphics[width=\textwidth]{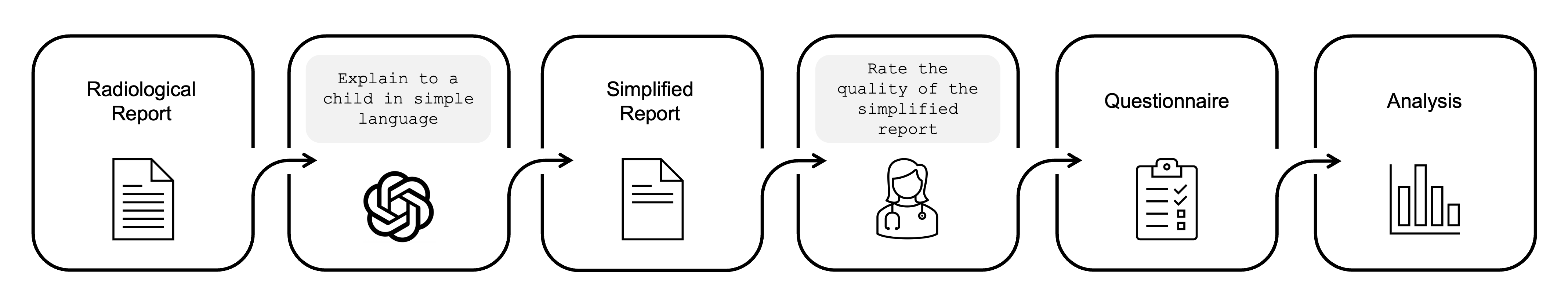}
    \caption{Summary of our evaluation pipeline. ChatGPT is prompted to generate multiple simplified reports. Radiologists are then tasked to assess their quality in a questionnaire. Lastly, the obtained data is evaluated.}
    \label{fig:flowchart}
\end{figure}

To investigate the capabilities of ChatGPT for simplification of radiology reports, we conducted an exploratory case study (Figure~\ref{fig:flowchart}) based on the research question: What is the radiologists' opinion on the quality of simplified radiology reports generated with ChatGPT?
We designed a questionnaire that included three fictitious original radiology reports written by an experienced radiologist and a unique simplified version of each generated by ChatGPT.
We asked 15 radiologists to rate the simplified reports and analyzed the results of the questionnaire.

\subsection{Original Radiology Reports}
\label{subsec:original_reports}

A radiologist with 10 years of experience wrote three artificial radiology reports.
The artificial reports did not contain any sensitive personalized information.
Each report contains multiple findings, which are associated with one another, e.g. tumor with edema, meniscus lesion with cruciate ligament lesion, and systemic metastasis.
All three reports are of moderate complexity, based on a definition, where a report is of low complexity in case it contains no pathological or at most one explicit pathological finding, and of high complexity in case of many and especially unclear findings.
The fictitious reports mimic real cases in clinical routine, i.e., the reports include previous medical information, describe the findings on the image, and contain a conclusion.

\begin{figure}[htbp]
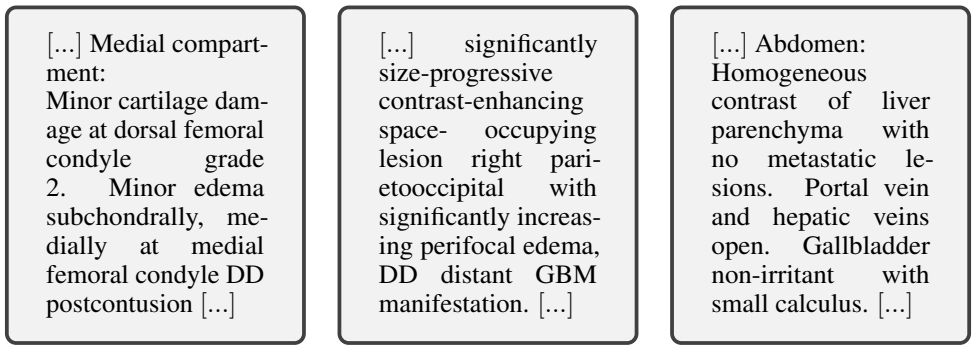

    \centering
    \begin{tabular}{lll}
     \begin{tcolorbox}[width=4cm]
     $\lbrack ... \rbrack$
     Medial compartment:
     
        Minor cartilage damage at dorsal femoral condyle grade 2. Minor edema subchondrally, medially at medial femoral condyle DD postcontusion
     $\lbrack ... \rbrack$
     \end{tcolorbox}  & 
     \begin{tcolorbox}[width=4cm]
          $\lbrack ... \rbrack$
         significantly size-progressive contrast-enhancing space-
occupying lesion right parietooccipital with significantly increasing perifocal edema, DD
distant GBM manifestation.
     $\lbrack ... \rbrack$
     \end{tcolorbox} &  
     \begin{tcolorbox}[width=4cm]
        $\lbrack ... \rbrack$
         Abdomen:
         
Homogeneous contrast of liver parenchyma with no metastatic lesions. Portal vein and hepatic veins open. Gallbladder non-irritant with small calculus.
     $\lbrack ... \rbrack$
     \end{tcolorbox}\\
    \end{tabular}
    \caption{Excerpts from the original radiology reports: \textit{Knee MRI} (left), \textit{Brain MRI} (middle), and \textit{Oncol. CT} (right). The full original reports are included in the Appendix.}
    \label{fig:orig-reports}
\end{figure}

The first report \textit{Knee MRI} (\ref{sec:appendix_original_report1}) describes a case in musculoskeletal radiology. 
The case of a neuroradiological MRI of the brain is the subject of the second report \textit{Head MRI} (\ref{sec:appendix_original_report2}).
It describes a follow-up examination and mentions comparisons to the previous examinations.
The third report (\ref{sec:appendix_original_report3}), referred to as \textit{Oncol. CT}, describes a fictitious oncological imaging event, reporting a follow-up whole-body CT scan.
To give an impression of the complex wordings in radiology reports, some excerpts of the original reports are shown in Figure~\ref{fig:orig-reports}.
\subsection{Simplifying Reports using ChatGPT}
\label{subsec:prompt}

The original radiology reports were simplified by prompting the ChatGPT (version December 15th, 2022) online interface with the request "Explain this medical report to a child using simple language:" followed by the original radiology report in text format.

This prompt was derived in a heuristic fashion using a separate fictitious radiology report that was different from the original reports described in Section \ref{subsec:original_reports}.
First, we tested different prompt versions of different lengths and compositions, including "Simplify", "Explain like I'm five", "Explain this medical report like I'm five", "Explain to a child using simple language", "Explain this medical report using simple language", and "Explain this medical report to a child using simple language".
Based on several trials, we gained the impression that the latter prompt produced the best and most stable results and we perceived the language of the generated reports to be significantly simpler compared to the original report.
We assume that this might be traced back to its detailed and explicit form, containing a concise call to action ("Explain"), specifying a narrow domain ("medical report"), and asking for a specific level of language complexity ("to a child using simple language").
During our tests, including phrases like "to a child" or "like I'm five" in the prompt on top of "simplify" or "using simple language" appeared to be beneficial for language simplification.
While our final prompt choice "Explain this medical report to a child using simple language" worked well for our radiology report examples in this study, we do not claim that this version is necessarily the best option for the task of simplifying radiology report with ChatGPT.

The output of ChatGPT is not deterministic by default. 
In principle, the model outputs probabilities for the next token in the autoregressive generation procedure.
However, the current settings of the ChatGPT interface do not allow tuning the temperature parameter responsible for handling the predicted token probabilities.
To account for the variability in the text output of ChatGPT and to achieve good coverage of its generative capability, we prompted the model 15 times for each of the three original reports, respectively, i.e., we generated 15 different simplified reports per original report.
Before each repeat of the prompt, we restarted the chat session to ensure that no tokens in the cache could distort the generated response.
All 45 ChatGPT-generated simplified reports can be found in Appendix \ref{appendix}.

\subsection{Questionnaire}

We designed a questionnaire to query radiologists on the quality of the simplified reports generated with ChatGPT.
We defined the quality of radiology reports as the combination of (i) \textit{factual correctness}, (ii) \textit{completeness}, and (iii) \textit{harmfulness}. 
\textit{Factual correctness} was not further specified. 
By our definition, a \textit{complete} report includes all key medical information, relevant to the patient.
We specify \textit{harm} as the potential that a subject could draw wrong conclusions from a simplification, which might result in any physical or psychological harm or lead to unwanted change in therapy or compliance.

The anonymized questionnaire "Questionnaire - Quality of Simplified Radiological Reports" (Appendix \ref{app:questionnaire_questions}) consisted of the three original radiology reports (section~\ref{subsec:original_reports}) and three unique simplified reports created with ChatGPT, respectively (section~\ref{subsec:prompt}). On the front page, the participating radiologists were explicitly informed that the simplified radiology reports were generated with "the machine learning language model ChatGPT" and received a description of the setup of the questionnaire and an instruction on how to answer.
Further, we asked for consent to participate and the years of experience, starting from the first year of residency.

Each questionnaire contained three blocks: (i) the original report, (ii) a single, randomly selected, unique simplified version of the original report, referred to as a "simplified report", and (iii) a series of questions to access the quality of the simplified reports.
We asked the participants to rate their level of agreement with each criterion on a five-point Likert scale (formulated as a statement, respectively).
Additionally, each question was accompanied by a follow-up question in which we asked the radiologists to provide text evidence for their assessment.
\begin{itemize}
    \item Factual Correctness: "The simplified radiological report is factually correct."\\
    Follow-up: "Highlight all incorrect text passages (if applicable) of the simplified report with a text marker".
    \item Completeness: "Relevant medical information for the patient is included in the simplified radiological report."\\
    Follow-up: "List all relevant medical information, which is missing in the simplified report (if applicable)."
    \item Potential Harm: "The simplified report leads patients to draw wrong conclusions, which might result in physical and/or psychological harm." \\
    Follow-up: "List all potentially harmful conclusions, which might be drawn from the simplified report (if applicable)."
\end{itemize}

15 radiologists with varying levels of experience from our clinic (Department of Radiology, University Hospital, LMU Munich) were asked to answer our questionnaire independently.
Each radiologist received the same three original reports and different, unique simplified reports.

\subsection{Evaluation}

The questionnaires were collected and checked for consent and completeness. 
For each participant, the years of experience were recorded, respectively.
The radiologists' ratings on the Likert Scales for factual correctness, completeness, and potential harm were evaluated for each of the three cases (\textit{Knee MRI}, \textit{Brain MRI}, \textit{Oncol. CT}).
Statistical parameters for the ordinal scales were calculated: median, 25\%-quantile ($Q_1$), 75\%-quantile ($Q_3$), interquartile range (IQR), minimum, maximum, mean, and standard deviation (SD).
The word count for the original reports and each simplified report was noted.
For each report, all passages of the highlighted text, as well as answers in the free-text fields were transcribed manually to a spreadsheet (Table~\ref{app:tab:questionnaire_answers}).
Additionally, the percentage of free-text questions and text highlights where text evidence was provided by the participants was calculated.
Finally, the free-text answers were inductively categorized by content.



\section{Results}
\label{sec:results}
In this section, we present our evaluation of the questionnaires, which were answered by 15 radiologists with a median (IQR) experience of 5 (9) years.
We analyzed their ratings on the Likert Scale, followed by the free-text analysis and the word count of the reports.

\subsection{Likert Scale Analysis}
\label{subsec:likert_scale_analysis}
\begin{figure}[h]
     \centering
     \begin{subfigure}[b]{\textwidth}
        \centering
        \includegraphics[width=0.7\textwidth]{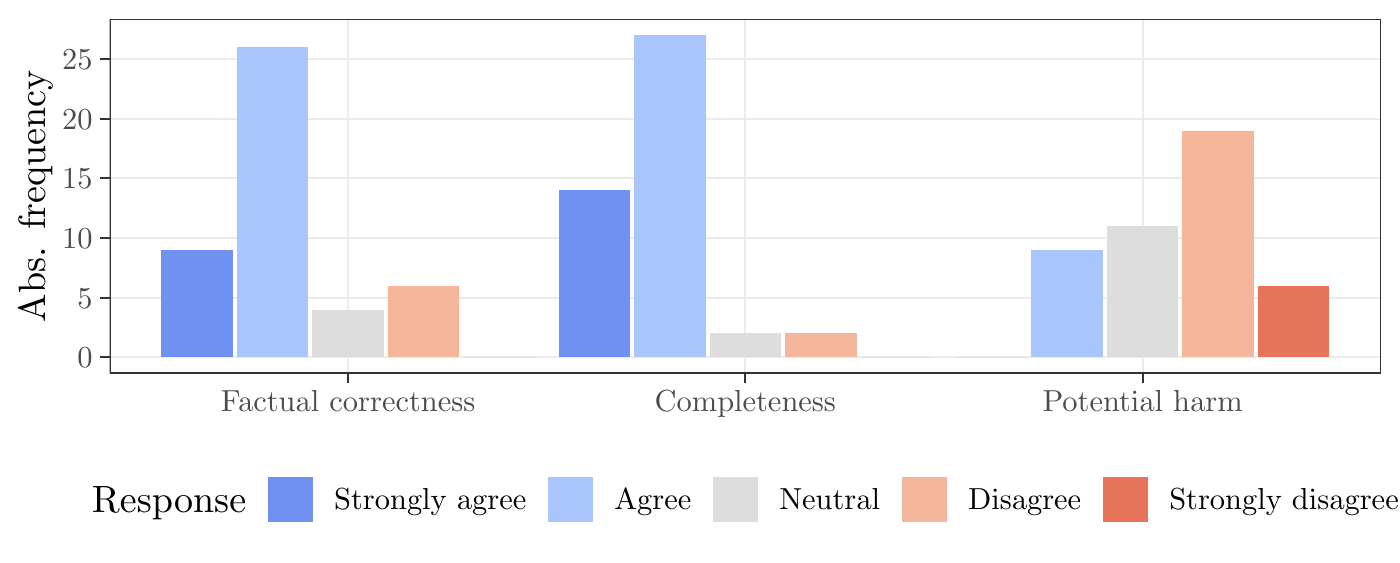}
        \caption{Results over all reports.}
    \label{fig:ratings-overall}
     \end{subfigure}
     \hfill
     \begin{subfigure}[b]{\textwidth}
        \centering
        \vspace{1mm}
        \includegraphics[width=\textwidth]{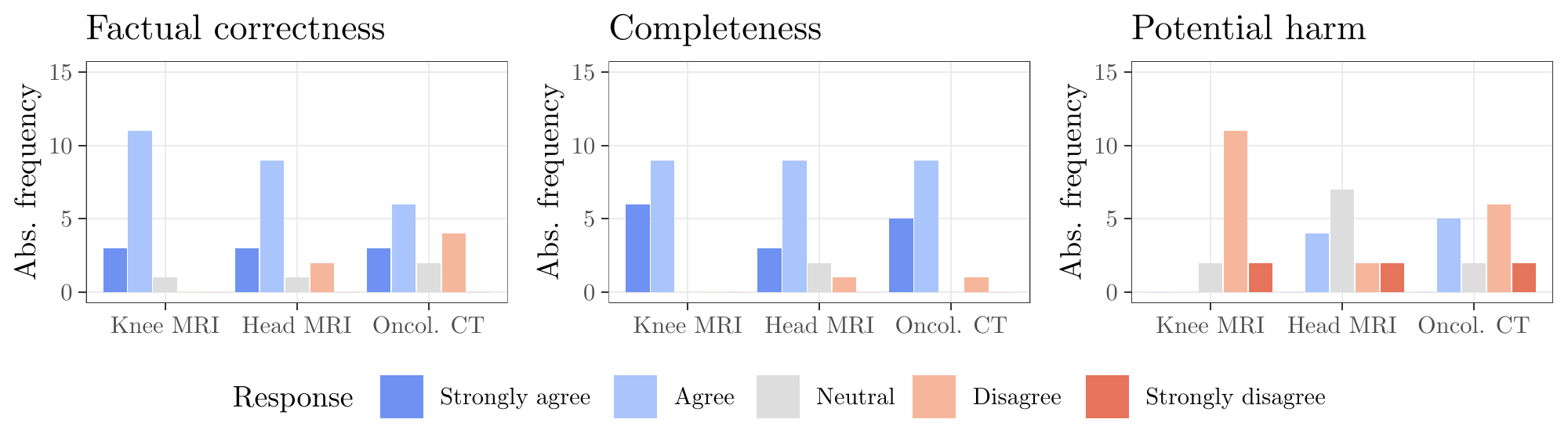}
        \caption{Results grouped by original report.}
        \label{fig:ratings-report}
     \end{subfigure}
        \caption{Frequency of radiologists’ ratings for all 45 simplified reports grouped by statement.}
        \label{fig:ratings}
\end{figure}

We first evaluated the radiologists' ratings for all 45 simplified reports (see Table~\ref{tab:sum_stat_total} and Figure~\ref{fig:ratings-overall}).
The participants generally agreed (median = 2) with the statements that the simplified reports are \textit{factually correct} and \textit{complete}, respectively.
For both quality criteria, 75\% of all ratings were given for "Agree" or "Fully agree" ($Q_3=2$), while "Fully disagree" was not selected at all.
In the case of \textit{completeness}, we found a slight tendency towards more positive ratings compared to \textit{factual correctness}.
25\% of all \textit{completeness} ratings were "Fully agree" ($Q_1=1$), more than for \textit{factual correctness}.
Further, "Neutral" and "Disagree" were only selected four times, but were selected ten times for \textit{factual correctness}.
Accordingly, the mean is slightly lower (closer to 1 = Fully agree) for \textit{completeness} (mean = 1.8) compared to \textit{factual correctness} (mean = 2.2).
In line with these findings, the participants disagreed (median = 4) on the potential of wrong conclusions drawn from the simplified reports resulting in physical and/or psychological harm.
Compared to \textit{factual correctness} (SD = 0.9) and \textit{completeness} (SD = 0.7), the radiologists' ratings of \textit{potential harm} are more broadly distributed (SD = 1.0) with a considerable proportion of ratings for "Neutral" and "Agree".
"Strongly Disagree" was not chosen at all.

In the second step, we also evaluated the radiologists' ratings for each of the three cases (\textit{Knee MRI}, \textit{Brain MRI}, and \textit{Oncol. CT}) individually. 
The results of the three reports are illustrated in Figure~\ref{fig:ratings-report} and summarized in Table~\ref{tab:sum_stat_by_rep}.
The median of the ratings showed no relevant differences in \textit{factual correctness}, \textit{completeness}, and \textit{potential harm}.
However, there is a tendency that the \textit{Head MRI} and the \textit{Oncol. CT} reports are judged to have a higher risk of patients drawing wrong conclusions with potentially harmful consequences.
\begin{table}[htbp]
\centering
\begin{tabular}{lcccccccc}
  \toprule
   \textbf{Question} & \textbf{Median} &  \textbf{$Q_1$} & \textbf{$Q_3$} & \textbf{IQR} & \textbf{Min} & \textbf{Max} &  \textbf{Mean} & \textbf{SD} \\ 
  \midrule
Factual correctness & 2 & 2 & 2 & 0 & 1 & 4 & 2.2 & 0.9 \\ 
Completeness & 2 & 1 & 2 & 1 & 1 & 4 & 1.8 & 0.7 \\ 
Potential harm & 4 & 3 & 4 & 1 & 2 & 5 & 3.5 & 1.0 \\ 
   \bottomrule
\end{tabular}
\vspace{2mm}
\caption{Summary statistics for \textit{factual correctness}, \textit{completeness}, and \textit{potential harm} over all 45 simplified reports. 1 = Strongly agree, 2 = Agree, 3 = Neutral, 4 = Disagree, 5 = Strongly disagree.} 
\label{tab:sum_stat_total}
\end{table}
\begin{table}[htbp]
\centering
\begin{tabular}{llcccccccc}
  \toprule
  \textbf{Question} & \textbf{Report} & \textbf{Median} &  \textbf{$Q_1$} & \textbf{$Q_3$} & \textbf{IQR} & \textbf{Min} & \textbf{Max} &  \textbf{Mean} & \textbf{SD} \\ 
  \midrule
  \multirow{3}*{Factual correctness} & Knee MRI & 2 & 2 & 2 & 0 & 1 & 3 & 1.9 & 0.5 \\ 
    & Head MRI & 2 & 2 & 2 & 0 & 1 & 4 & 2.1 & 0.9 \\ 
     & Oncol. CT & 2 & 2 & 3.5 & 1.5 & 1 & 4 & 2.5 & 1.1 \\
    \midrule
  \multirow{3}*{Completeness} & Knee MRI & 2 & 1 & 2 & 1 & 1 & 2 & 1.6 & 0.5 \\ 
   & Head MRI & 2 & 2 & 2 & 0 & 1 & 4 & 2.1 & 0.8 \\ 
   & Oncol. CT & 2 & 1 & 2 & 1 & 1 & 4 & 1.8 & 0.8 \\ 
      \midrule
  \multirow{3}*{Potential harm} & Knee MRI & 4 & 4 & 4 & 0 & 3 & 5 & 4 & 0.5 \\ 
   & Head MRI & 3 & 2.5 & 3.5 & 1 & 2 & 5 & 3.1 & 1 \\ 
  & Oncol. CT & 4 & 2 & 4 & 2 & 2 & 5 & 3.3 & 1.1 \\ 
   \bottomrule
\end{tabular}
\vspace{2mm}
\caption{Summary statistics for \textit{factual correctness}, \textit{completeness}, and \textit{potential harm} grouped by original report. 1 = Strongly agree, 2 = Agree, 3 = Neutral, 4 = Disagree, 5 = Strongly disagree.}
\label{tab:sum_stat_by_rep}
\end{table}

\subsection{Free-text Analysis}
In the following, we summarize the results of the follow-up questions.
For each follow-up question, we applied inductive categorization of the free-text data by content.
We present the identified categories along with a few selected examples for each.
51\% of the participants highlighted incorrect passages, while only 22\% and 36\% of the radiologists listed missing relevant information and potentially harmful conclusions, respectively (Table \ref{tab:perc_answers}). 
\begin{table}[b]
\centering
\begin{tabular}{lc}
  \toprule
  & \textbf{Answered (\%)} \\ 
  \midrule
  Highlight incorrect passages & 51 \\ 
  List missing relevant information & 22 \\ 
  List harmful conclusions & 36 \\ 
   \bottomrule
\end{tabular}
\vspace{2mm}
\caption{Percentage of free-text questions and text highlights where text evidence was provided by the participants.}
\label{tab:perc_answers}
\end{table}

\subsubsection{Highlights of Incorrect Text Passages}
\label{subsec:highlights}

We asked the participants to mark incorrect passages in the text (see highlighted passages in the simplified texts in Appendix~\ref{appendix}), which unveiled a variety of weak points in the simplified reports of ChatGPT.

\paragraph{Misinterpretation of medical terms}
We observed that in the simplified text generated by ChatGPT medical terminology was misinterpreted in several cases.
The abbreviation "DD" (differential diagnosis) was often mistaken as a final diagnosis (\ref{app:rep2.6}, \ref{app:rep2.8}), e.g., the simplified report contains "The conclusion of the report is that the mass on the right side of the head is a type of cancer called a glioblastoma $\lbrack ... \rbrack$" for the original statement "$\lbrack ... \rbrack$ DD distant GBM manifestation".
Furthermore, "thyroid struma" was described by ChatGPT in one instance as "infection in their thyroid gland" (\ref{app:rep3.13}) and in another as "extra thyroid gland" (\ref{app:rep3.7}).
The finding "No pneumothorax" was described as "no holes in the walls between the lungs and chest" (\ref{app:rep3.3}).
The lateral compartment of the knee was described as being on the "part of your body that's on the outside" (\ref{app:rep1.15}).
The term "retroperitoneal" (referring to lymph nodes in need of further control) was wrongly interpreted as "in the person’s back" (\ref{app:rep3.9}).
Lastly, a growing mass of "currently max. 22\,mm" was incorrectly simplified as a "small, abnormal growth $\lbrack ... \rbrack$ that has gotten bigger" (\ref{app:rep2.3}).

\paragraph{Imprecise language}
In some cases we observed that the automated simplification led to passages of imprecise language: 
The medial compartment of the knee was described as the "middle part of your leg" (\ref{app:rep1.5}), the brain as the "head" (\ref{app:rep2.7}), the abdomen as the "stomach" (\ref{app:rep3.7}) and thyroid struma as "thyroid problems" (\ref{app:rep3.12}).
Partial regression was described as "gotten smaller and is not spreading as much as before" (\ref{app:rep3.12}), intercondylar area as "the area between the two bones that make up the lower part of the leg" (\ref{app:rep1.4}) and cartilage was described as "hard, smooth substance" (\ref{app:rep1.9}).
Metastases were imprecisely simplified as "spots" (\ref{app:rep3.5}).
A case of ambiguous simplification of the original text is found where the simple report ended with "$\lbrack ... \rbrack$ is no evidence of the cancer spreading to other parts of the body." (\ref{app:rep3.13}). In the original report pulmonary metastases were described.

\paragraph{Hallucination}
Some findings were not contained in the original report, i.e. they were made up in the simplified reports, for instance,   
"no signs of cancer $\lbrack ... \rbrack$ in the thyroid gland" (\ref{app:rep3.3}) or "brain does not seem to be damaged" (\ref{app:rep2.9}). 

\paragraph{Odd language}
We observed odd and unsuited language in several simplified reports.
The idiom "wear and tear" (\ref{app:rep1.4}) was used to describe the degeneration of tendons and a radiological examination was circumscribed as "to see how the cancer is doing" (\ref{app:rep3.8}).
In report \ref{app:rep2.6} contrast media was paraphrased with the word "dye" and tumor as a "group of abnormal cells". 

\paragraph{Grammatical errors}
Grammatical errors such as "CT scan" instead of "CT scanner" (\ref{app:rep3.7}) were the exception. 

\subsubsection{Missing Key Medical Information}
\label{subsec:missing_info}

Furthermore, the simplified reports often missed key medical findings mentioned in the original report.

\paragraph{Missed findings}
For \textit{Knee MRI} (\ref{sec:appendix_original_report1}), a participant noted that the information on cartilage damage on the inner side of the knee was missing in the simplified report (\ref{app:rep1.5}).
In the \textit{Head MRI} report (\ref{sec:appendix_original_report2}) the information on "no intermediate or recent ischemia" (\ref{app:rep2.2}) was missing in one instance. 
Another missed finding was the growth of the lesion on the parietooccipital side in the conclusion of the simplified report (\ref{app:rep2.4}).
For \textit{Oncol. CT} (\ref{sec:appendix_original_report3}) one participant reported that mentioning "spots" in contrast to metastases removes medical information.
Furthermore the fact, that the solid portions of the pulmonary metastases are decreasing (\ref{app:rep3.5}) - consistent with therapy response - is not included in the simplified report.
In another simplified report (\ref{app:rep3.13}), where the thyroid struma is misinterpreted as infection, the description of the enlargement is missing.

\paragraph{Unspecific location}
We observed missing or unspecific location information in the simplified reports.
In the simplified report \ref{app:rep1.15} the location of the described finding (knee) was not mentioned.
A participant mentioned that  "the back" (\ref{app:rep3.14}) is not precise enough to describe the affected retroperitoneum.
Non-specific descriptions of the location resulted in problems with the identification of the present tumor and the excised tumor (\ref{app:rep2.11}).

\subsubsection{Potentially Harmful}
\label{subsubsec:potentially_harmful}

Finally, we asked the participants to list all potentially harmful conclusions, which might be drawn from the simplified report.
The given answers corresponded mostly to already highlighted passages of incorrect text or missing key medical information and corroborated the detected weaknesses.
Odd language and grammatical errors were not considered potentially harmful.

\paragraph{Misinterpretation of medical terms}
Participants listed the misinterpretation of differential diagnosis ("DD") for final diagnosis as potentially harmful for the patient, e.g., "GBM is one (likely) DD which implies that other DDs exist (e.g. radionecrosis)" (\ref{app:rep2.7}).
Lymph nodes were presented in the simplified report like "they might have cancer"(\ref{app:rep3.9}).
This was considered harmful since the original report stated that there was "no evidence of recurrence or new lymph node metastases".
Furthermore, the wording "small growth" (\ref{app:rep2.3}) was deemed a harmful conclusion, as the original report \textit{Head MRI} (\ref{sec:appendix_original_report2}) states a progression in size, which "almost doubled".

\paragraph{Hallucination}
Participants assessed the hallucination "brain does not seem to be damaged" (\ref{app:rep2.9}) as potentially harmful to the patient, as the original report describes a growing mass.

\paragraph{Imprecise language}
We observed that imprecise language highlighted as incorrect was sometimes also graded as potentially harmful to the patient.
One participant noted that "there is no (new) spread of cancer" in the simplified oncological report (\ref{app:rep3.13}).
Additionally, participants listed these potentially harmful conclusions: "not clear that spots are pulmonal metastasis" (\ref{app:rep3.5}), "changes happened recently (how recently?)", and "some extra fluid $\neq$ sign. incr. edema" (\ref{app:rep2.13}).

\paragraph{Missed findings}
We found that potentially harmful conclusions can originate from missing information in the simplified report.
The sentence "the parts that are staying the same size are changing" (\ref{app:rep3.5}) was found to be an unclear statement missing interpretation.
One participant stated that "information $\lbrack ... \rbrack$ about the lesion showing growth" is missing from the conclusion of the simplified report (\ref{app:rep2.4}).

\paragraph{Unspecific location}
Cases, where ChatGPT misses the correct information about a location of a disease, can lead to potentially harmful conclusions.
Examples were: "misunderstanding which lesion is stable and which one is in progress can lead the patient to some wrong expectations" (\ref{app:rep2.11}) and "patient might be misled to look at cancer spots on their back" (\ref{app:rep3.14}).

\subsection{Word Count}
The median word count for the simplified \textit{Head MRI} and \textit{Oncol. CT} reports were comparable to their respective original reports (Table~\ref{tab:word_count}).
The simplified \textit{Knee MRI} report with a median (IQR) word count of 414 (60) is generally longer compared to the original report (word count = 222).
\begin{table}[htbp]
\centering
\begin{tabular}{lccccc}
  \toprule
   \textbf{Report} &  \textbf{Original} & \multicolumn{4}{c}{\textbf{Simplified}} \\ \cmidrule{3-6}
 &  & Median & Min & Max & IQR \\ 
  \midrule
  Knee MRI & 222 & 414 & 249 & 538  & 60  \\ 
  Head MRI & 117 & 150 & 91 & 241  & 79  \\ 
  Oncol. CT  & 241 & 249 &  111 & 386  & 94  \\ 
   \bottomrule
\end{tabular}
\vspace{2mm}
\caption{Word count of the original radiology reports and the ChatGPT-generated simplified reports.} 
\label{tab:word_count}
\end{table}
%

\section{Discussion}

In this exploratory study, we evaluated the opinion of radiologists on the quality of radiology reports which were simplified by ChatGPT.
Participating radiologists rated the simplified radiology reports according to (i) \textit{factual correctness}, (ii) \textit{completeness}, and (iii) \textit{potential harmfulness}.
In the following, we discuss the implication of our results presented in Section \ref{sec:results}.

\subsection{Quality of Simplified Reports Generated with ChatGPT}
We prompted ChatGPT to generate 45 simplified radiology reports for this study. 
According to the authors of ChatGPT, the model is in principle able to state to not know the answer \cite{chatgpt}.
However, in no instance, the model refused to give an answer or indicated that the response could be incorrect while generating consistent plausible-sounding outputs.

\subsubsection{Factual Correctness}
Although ChatGPT was not explicitly trained for the purpose of simplifying radiology reports, most radiologists agreed that the generated simplified reports are factually correct (median = 2, Section \ref{sec:results}).
However, some radiologists also identified statements that were not considered factually correct (Figure \ref{fig:ratings}).
In particular, we identified the following error categories: \textit{misinterpretation of medical terms}, \textit{imprecise language}, \textit{hallucination}, \textit{odd language}, and \textit{grammatical errors}.

On the one hand, we speculate that the problem of \textit{misinterpretation of medical terms}, \textit{imprecise language}, \textit{odd language}, and \textit{grammatical errors} may be solved by the adoption of a ChatGPT-like model for the medical domain, as well as by optimizing the prompt.
On the other hand, we also observed \textit{hallucination}, which is a type of behavior where the model adds statements to the simplified report that cannot be derived from the original report (Section \ref{subsec:highlights}).
This is an intrinsic problem in generative models like LLMs \cite{DBLP:conf/acl/ZhouNGDGZG21}.

\subsubsection{Completeness}
Prompting ChatGPT for simplification is likely to cause responses, where content is removed or rephrased in oversimplified terms.
Therefore, we tasked radiologists to verify the \textit{completeness} of the simplified reports.
Overall, the radiologists agreed that the simplified reports contained the medical information relevant for the patients (median = 2, Section \ref{subsec:likert_scale_analysis}).
This indicates that ChatGPT is able to identify the most important aspects of the complex medical content of radiology reports.
Nevertheless, the participants also detected missing key medical information, which we summarized in the two categories \textit{missed findings} and \textit{unspecific location} (Section \ref{sec:results}).
These findings suggest that paraphrasing and explaining complex medical terminology in more simple terms can have the side effect of losing medical context and professional preciseness. 

It is an open question if the problem of missing key medical information could be solved by adapting the prompt or if this is an unreachable goal for the current version of ChatGPT.

\subsubsection{Potential Harm}
Non-maleficence, i.e., to not harm individual patients, is a major principle of medical ethics, also in the context of AI applications \cite{beauchamp_principles_2019,hagendorff_ethics_2020}.
Thus, we asked the radiologists if they think that "The simplified report leads patients to draw wrong conclusions, which might result in physical and/or psychological harm.” 

Overall, radiologists disagreed that potentially harmful conclusions will be drawn from the ChatGPT-generated reports (median = 4, Table~\ref{tab:sum_stat_total}).
This finding may encourage patients to use ChatGPT or similar LLMs as a tool to simplify radiology reports.
However, as shown in Section~\ref{subsec:highlights} and \ref{subsec:missing_info}, simplified reports can contain errors and key medical findings might be missed.
About half of the radiologists found critical statements in the simplified reports with a high potential of leading patients to wrong conclusions, which bears the great risk of causing physical or psychological harm (Section~\ref{subsubsec:potentially_harmful}).
For example, the medical finding that certain lymph nodes "might have cancer" is an overstatement in the simplified report \ref{app:rep3.9}. The original report clearly states that there is "No evidence of recurrence or new lymph node metastases" and only suggests to "further control" the "Unchanged prominent retroperitoneal lymph nodes".  
This might cause psychological harm since the patient could conclude that the CT shows hints of new lymph node metastases.
Another example can be found in the simplified report \ref{app:rep2.3}. Here, only a "small growth" of the contrast-enhancing mass is described, whereas the original report (\ref{sec:appendix_original_report2}) states a progression in size, which "almost doubled". This is a clear understatement, which could lead the patient to underestimate the progression of the disease.

While, overall, radiologists do not see a high risk for potential harm, they still point out that single errors occur which might be potentially harmful.
Users of ChatGPT should be aware of this problem.

\subsection{Opportunities and Challenges}
With openly available LLMs like ChatGPT, patients might use these tools to simplify their radiology reports in the near future.
This scenario raises a number of opportunities and challenges.

\paragraph{Patient-centered care}
Improved accessibility of personal medical information  may be beneficial to the patient in multiple ways.
Simplified reports give patients the chance to better comprehend their health situation instead of making them feel overwhelmed by the medical terminology found in the original report and might help patients to better prepare themselves for future doctor-patient interactions.
For instance, the simplified report might raise questions patients could then ask the medical practitioner.
This might lead to an improved understanding of their own health condition which, in turn, empowers the patients, strengthens their position throughout the medical treatment, and facilitates informed patient decisions.
Simplified radiology reports are particularly valuable for patients who do not speak the native language of the country in which they are treated.
Even though ChatGPT works best for English, other languages are supported with varying quality.
In the case of poorly supported languages, reports could be translated into English using established tools prior to simplification.
Overall, easily accessible simplified radiology reports may help to increase patient autonomy and allow them to take on a more active role throughout their medical treatment process facilitating patient-centered care.

However, when patients simplify their reports in advance of a medical consultation, patients will not be assisted by an expert while reading and attempting to interpret them.
Without medical advice, patients could misinterpret certain findings or even learn about a life-threatening diagnosis from their simplified report, but the information might not be accompanied by additional context, e.g., therapy options.
This could put the patient in a state of mental stress.
Additionally, there might be an increased risk of patients making their own clinical decisions, such as delaying or even omitting further doctor appointments or terminating a therapy without professional medical consultation.  
Even though the language conveying the information might be simplified, the content of radiology reports and the process of clinical decision-making remains complex.

\paragraph{Zero-shot simplification of radiology reports}
In this study, we found that ChatGPT already performs surprisingly well at simplifying radiology reports out of the box.
It produces fairly accurate and complete simplified reports.
If this technology continues to become even more easily accessible for laypersons, it has the potential to be applied by a wide range of users in the near future.
To fully leverage this tool as a radiology report simplifier with maximum benefit for patients, several technical hurdles need to be overcome. 

Among others, we see the following challenges:
ChatGPT was not developed for the specific task of simplifying radiology reports.
A domain adaption of ChatGPT to the medical field could further improve the quality of generated simplified radiology reports, e.g., leading to a more correct interpretation of medical terms.
The task of simplification could be enforced by using a reward model trained to estimate the quality of simplified radiology reports in the RLHF training procedure of ChatGPT.
Like other LLMs, ChatGPT might have intrinsic biases due to imbalanced training data \cite{bommasani2021opportunities, brown2020language}.
We hypothesize that for the task of simplifying radiology reports, rare pathologies might be handled less accurately than more common ones.
Furthermore, medicine is an active and dynamic field of science, which is subject to progress and change.
However, ChatGPT - like most LLMs - is trained on data with a certain timestamp \cite{bender2021dangers}.
New research findings in medicine are thus not reflected in the model's output.
Another problematic property of ChatGPT is the non-deterministic output.
ChatGPT generates text by predicting tokens on a probabilistic basis, which does not guarantee the most likely response.
While this fosters dynamic dialogue, it could also lead to high variance with respect to output reliability.
As of now, there is no way for users to control ChatGPT, e.g., with a temperature parameter or seed.
This makes it difficult, if not almost impossible, to reproduce a certain output, even when applying the same input multiple times.
Finally, there are privacy concerns when sensitive data is uploaded to a proprietary service.

\paragraph{Potential future clinical application}
Simplified radiology reports have the potential to increase patient autonomy during the treatment process.
With the release of ChatGPT, patients now have unprecedented possibilities to automatically generate such simplified reports themselves.
However, there is an unavoidable risk for patients to draw potentially harmful conclusions from those reports, when no medical oversight is provided.

We envision a future integration of ChatGPT-like LLMs directly in the clinic or radiology centers.
In this scenario, a simplified radiology report would always be automatically generated with an LLM alongside the original report, proofread by a radiologist, and corrected where necessary.
Both reports would then be issued to the patient.
We imagine the application of an appropriate LLM that is adapted to the medical domain, properly certified, and hosted in a privacy-preserving way, e.g., on-premise.
In our opinion, this would be a cost-efficient way to leverage state-of-the-art technology for improving patient-centered care.

\subsection{Limitations of the Study}

The number of original radiology reports (n = 3) and experts (n = 15) for assessing the quality of the simplified reports is small.
Further, we used fictitious radiology reports written by an experienced radiologist instead of original reports of patients to protect patient privacy.
The radiologist who generated the simplified reports is German, i.e., not a native English speaker.
A German abbreviation ("Z.n." for "Zustand nach", which can be paraphrased as "following"), which does not exist in English, was found in the original reports.
The original reports used in this study are at a medium complexity level.
It is unclear how the quality of less or more complex reports would be rated.
Also, the language simplicity of ChatGPT-generated radiology reports was analyzed only qualitatively during the design process of the input prompt ("Explain this medical report to a child using simple language").
We did not further investigate the ability of ChatGPT to simplify language with respect to established guidelines for simple language.
We do not claim that our prompt for triggering a simplification of radiology reports through ChatGPT, is optimal for this task. Our prompt is the result of a heuristic evaluation of experimenting with different inputs.
Finally, the quality of the simplified reports generated with ChatGPT  was only evaluated from a radiologist's point of view.
The patients' perspective on the readability, accessibility, and added value of those simplified reports was not part of this study.


\section{Conclusion}

In this case study, we asked radiologists to assess the quality of simplified radiology reports generated with ChatGPT.
We found that most radiologists agree that the simplified reports are factually correct, complete, and not potentially harmful to patients.
This positive vote illustrates the great potential that LLMs could have for medical text simplification.
In particular, we argue that simplified reports generated with LLMs like ChatGPT would strengthen patients' autonomy.
However, the radiologists also identified factual incorrect statements, missing key medical information, and text passages, which might lead patients to draw potentially harmful conclusions from the simplified reports, if physicians are not kept in the loop.
We hypothesize that LLMs like ChatGPT may be used by patients to simplify radiology reports that contain critical medical information, even though the models were not developed for this specific medical use case.
Thus, we see a need for technical improvements and fine-tuning of ChatGPT or similar LLMs to the specific task of simplifying radiology reports. 

Our exploratory case study can only give first insights into the field of automated simplification of radiology reports with ChatGPT, especially due to the limited number of participants and radiology report cases. 
Given the great potential of LLMs for simplifying radiology reports, there is a demand for future research to validate the findings of our case study and to further explore the many possibilities of this new technology in the medical domain.
This requires quantitative studies on the quality of simplified radiology reports generated with ChatGPT and similar LLMs.
Also, the opinion of patients regarding the accessibility of those simplified reports and their overall added value within the treatment process should be investigated.
We propose to include the automatic generation of simplified radiology reports in the clinical process based on a domain-adapted model.
These should be approved by experts and issued to patients alongside their original report.
Overall, we see great potential in using LLMs like ChatGPT to improve patient-centered care in radiology and other medical domains.

\section*{Acknowledgments}
We thank all radiologists who participated in this study.
This work has been partially funded by the Deutsche Forschungsgemeinschaft (DFG, German Research Foundation) as part of BERD@NFDI - grant number 460037581.


\printbibliography

@article{vaswani2017attention,
  title={Attention is all you need},
  author={Vaswani, Ashish and Shazeer, Noam and Parmar, Niki and Uszkoreit, Jakob and Jones, Llion and Gomez, Aidan N and Kaiser, {\L}ukasz and Polosukhin, Illia},
  journal={Advances in neural information processing systems},
  volume={30},
  year={2017}
}

@article{christiano2017deep,
  title={Deep reinforcement learning from human preferences},
  author={Christiano, Paul F and Leike, Jan and Brown, Tom and Martic, Miljan and Legg, Shane and Amodei, Dario},
  journal={Advances in neural information processing systems},
  volume={30},
  year={2017}
}

@article{ouyang2022training,
  title={Training language models to follow instructions with human feedback},
  author={Ouyang, Long and Wu, Jeff and Jiang, Xu and Almeida, Diogo and Wainwright, Carroll L and Mishkin, Pamela and Zhang, Chong and Agarwal, Sandhini and Slama, Katarina and Ray, Alex and others},
  journal={arXiv preprint arXiv:2203.02155},
  year={2022}
}

@article{bommasani2021opportunities,
  title={On the opportunities and risks of foundation models},
  author={Bommasani, Rishi and Hudson, Drew A and Adeli, Ehsan and Altman, Russ and Arora, Simran and von Arx, Sydney and Bernstein, Michael S and Bohg, Jeannette and Bosselut, Antoine and Brunskill, Emma and others},
  journal={arXiv preprint arXiv:2108.07258},
  year={2021}
}

@article{brown2020language,
  title={Language models are few-shot learners},
  author={Brown, Tom and Mann, Benjamin and Ryder, Nick and Subbiah, Melanie and Kaplan, Jared D and Dhariwal, Prafulla and Neelakantan, Arvind and Shyam, Pranav and Sastry, Girish and Askell, Amanda and others},
  journal={Advances in neural information processing systems},
  volume={33},
  pages={1877--1901},
  year={2020}
}

@article{schulman2017proximal,
  title={Proximal policy optimization algorithms},
  author={Schulman, John and Wolski, Filip and Dhariwal, Prafulla and Radford, Alec and Klimov, Oleg},
  journal={arXiv preprint arXiv:1707.06347},
  year={2017}
}

@inproceedings{devlin2018bert,
  author    = {Jacob Devlin and
               Ming{-}Wei Chang and
               Kenton Lee and
               Kristina Toutanova},
  title     = {{BERT:} Pre-training of Deep Bidirectional Transformers for Language
               Understanding},
  booktitle = {Proceedings of the 2019 Conference of the North American Chapter of the Association for Computational Linguistics: Human Language Technologies, {NAACL-HLT}},
  pages     = {4171--4186},
  year      = {2019},
  url       = {https://doi.org/10.18653/v1/n19-1423},
  doi       = {10.18653/v1/n19-1423},
  bibsource = {dblp computer science bibliography, https://dblp.org}
}

@article{shardlow2014survey,
  title={A survey of automated text simplification},
  author={Shardlow, Matthew},
  journal={International Journal of Advanced Computer Science and Applications},
  volume={4},
  number={1},
  pages={58--70},
  year={2014}
}

@article{al2021automated,
  title={Automated text simplification: a survey},
  author={Al-Thanyyan, Suha S and Azmi, Aqil M},
  journal={ACM Computing Surveys (CSUR)},
  volume={54},
  number={2},
  pages={1--36},
  year={2021},
  publisher={ACM New York, NY, USA}
}

@inproceedings{kvist2013professional,
  title={Professional language in swedish radiology reports--characterization for patient-adapted text simplification},
  author={Kvist, Maria and Velupillai, Sumithra},
  booktitle={Scandinavian Conference on Health Informatics},
  pages={55--59},
  year={2013},
  organization={Link{\"o}ping University Electronic Press}
}

@inproceedings{abrahamsson2014medical,
  title={Medical text simplification using synonym replacement: Adapting assessment of word difficulty to a compounding language},
  author={Abrahamsson, Emil and Forni, Timothy and Skeppstedt, Maria and Kvist, Maria},
  booktitle={Proceedings of the 3rd Workshop on Predicting and Improving Text Readability for Target Reader Populations (PITR)},
  pages={57--65},
  year={2014}
}

@article{agarwal2022explain,
  title={Explain to me like I am five--Sentence Simplification Using Transformers},
  author={Agarwal, Aman},
  journal={arXiv preprint arXiv:2212.04595},
  year={2022}
}

@article{qenam2017text,
  title={Text simplification using consumer health vocabulary to generate patient-centered radiology reporting: translation and evaluation},
  author={Qenam, Basel and Kim, Tae Youn and Carroll, Mark J and Hogarth, Michael and others},
  journal={Journal of medical Internet research},
  volume={19},
  number={12},
  year={2017},
  publisher={JMIR Publications Inc., Toronto, Canada}
}

@inproceedings{ramadier2017radiological,
  title={Radiological Text Simplification Using a General Knowledge Base},
  author={Ramadier, Lionel and Lafourcade, Mathieu},
  booktitle={International Conference on Computational Linguistics and Intelligent Text Processing},
  pages={617--627},
  year={2017},
  organization={Springer}
}

@inproceedings{zhang2018learning,
  author    = {Yuhao Zhang and
               Daisy Yi Ding and
               Tianpei Qian and
               Christopher D. Manning and
               Curtis P. Langlotz},
  title     = {Learning to Summarize Radiology Findings},
  booktitle = {Proceedings of the Ninth International Workshop on Health Text Mining
               and Information Analysis, Louhi@EMNLP},
  pages     = {204--213},
  publisher = {Association for Computational Linguistics},
  year      = {2018},
  url       = {https://doi.org/10.18653/v1/w18-5623},
  doi       = {10.18653/v1/w18-5623},
  bibsource = {dblp computer science bibliography, https://dblp.org}
}

@article{cai2021chestxraybert,
  title={ChestXRayBERT: A Pretrained Language Model for Chest Radiology Report Summarization},
  author={Cai, Xiaoyan and Liu, Sen and Han, Junwei and Yang, Libin and Liu, Zhenguo and Liu, Tianming},
  journal={IEEE Transactions on Multimedia},
  year={2021},
  publisher={IEEE}
}

@inproceedings{liang2022fine,
  title={Fine-tuning BERT Models for Summarizing German Radiology Findings},
  author={Liang, Siting and Kades, Klaus and Fink, Matthias and Full, Peter and Weber, Tim and Kleesiek, Jens and Strube, Michael and Maier-Hein, Klaus},
  booktitle={Proceedings of the 4th Clinical Natural Language Processing Workshop},
  pages={30--40},
  year={2022}
}

@article{barrett_patient-centered_2021,
        title = {Patient-centered {Reporting} in {Radiology}: {A} {Single}-site {Survey} {Study} of {Lung} {Cancer} {Screening} {Results}},
        volume = {36},
        issn = {0883-5993},
        shorttitle = {Patient-centered {Reporting} in {Radiology}},
        url = {http://journals.lww.com/thoracicimaging/Abstract/2021/11000/Patient_centered_Reporting_in_Radiology__A.5.aspx},
        doi = {10.1097/RTI.0000000000000591},
        number = {6},
        urldate = {2022-12-21},
        journal = {Journal of Thoracic Imaging},
        author = {Barrett, Spencer K. and Patrie, James and Kitts, Andrea B. and Hanley, Michael and Swanson, Christina M. and Vitzthum von Eckstaedt, Hans and Krishnaraj, Arun},
        month = nov,
        year = {2021},
        pages = {367},
}

@article{martin-carreras_readability_2019,
        title = {Readability of radiology reports: implications for patient-centered care},
        volume = {54},
        issn = {1873-4499},
        shorttitle = {Readability of radiology reports},
        doi = {10.1016/j.clinimag.2018.12.006},
        language = {eng},
        journal = {Clinical Imaging},
        author = {Martin-Carreras, Teresa and Cook, Tessa S. and Kahn, Charles E.},
        year = {2019},
        pmid = {30639521},
        pages = {116--120},
}

@inproceedings{macavaney2019ontology,
  title={Ontology-aware clinical abstractive summarization},
  author={MacAvaney, Sean and Sotudeh, Sajad and Cohan, Arman and Goharian, Nazli and Talati, Ish and Filice, Ross W},
  booktitle={Proceedings of the 42nd International ACM SIGIR Conference on Research and Development in Information Retrieval},
  pages={1013--1016},
  year={2019}
}

@inproceedings{gigioli2018domain,
  title={Domain-aware abstractive text summarization for medical documents},
  author={Gigioli, Paul and Sagar, Nikhita and Rao, Anand and Voyles, Joseph},
  booktitle={2018 IEEE International Conference on Bioinformatics and Biomedicine (BIBM)},
  pages={2338--2343},
  year={2018},
  organization={IEEE}
}

@article{du2020biomedical,
  title={Biomedical-domain pre-trained language model for extractive summarization},
  author={Du, Yongping and Li, Qingxiao and Wang, Lulin and He, Yanqing},
  journal={Knowledge-Based Systems},
  volume={199},
  year={2020},
  publisher={Elsevier}
}

@article{lee2020cerc,
  title={CERC: an interactive content extraction, recognition, and construction tool for clinical and biomedical text},
  author={Lee, Eva K and Uppal, Karan},
  journal={BMC Medical Informatics and Decision Making},
  volume={20},
  number={14},
  pages={1--14},
  year={2020},
  publisher={BioMed Central}
}

@article{chaves2022automatic,
  title={Automatic Text Summarization of Biomedical Text Data: A Systematic Review},
  author={Chaves, Andrea and Kesiku, Cyrille and Garcia-Zapirain, Begonya},
  journal={Information},
  volume={13},
  number={8},
  pages={393},
  year={2022},
  publisher={MDPI}
}

@article{yi_readability_2019,
        title = {Readability of {Lumbar} {Spine} {MRI} {Reports}: {Will} {Patients} {Understand}?},
        volume = {212},
        issn = {0361-803X},
        shorttitle = {Readability of {Lumbar} {Spine} {MRI} {Reports}},
        url = {http://www.ajronline.org/doi/10.2214/AJR.18.20197},
        doi = {10.2214/AJR.18.20197},
        number = {3},
        urldate = {2022-12-21},
        journal = {American Journal of Roentgenology},
        author = {Yi, Paul Hyunsoo and Golden, Sean Kenney and Harringa, John B. and Kliewer, Mark A.},
        month = mar,
        year = {2019},
        note = {Publisher: American Roentgen Ray Society},
        pages = {602--606},
}

@article{shoeybi2019megatron,
  title={Megatron-lm: Training multi-billion parameter language models using model parallelism},
  author={Shoeybi, Mohammad and Patwary, Mostofa and Puri, Raul and LeGresley, Patrick and Casper, Jared and Catanzaro, Bryan},
  journal={arXiv preprint arXiv:1909.08053},
  year={2019}
}

@article{raffel2020exploring,
  title={Exploring the limits of transfer learning with a unified text-to-text transformer.},
  author={Raffel, Colin and Shazeer, Noam and Roberts, Adam and Lee, Katherine and Narang, Sharan and Matena, Michael and Zhou, Yanqi and Li, Wei and Liu, Peter J and others},
  journal={J. Mach. Learn. Res.},
  volume={21},
  number={140},
  pages={1--67},
  year={2020}
}

@article{scao2022bloom,
  title={Bloom: A 176b-parameter open-access multilingual language model},
  author={Scao, Teven Le and Fan, Angela and Akiki, Christopher and Pavlick, Ellie and Ili{\'c}, Suzana and Hesslow, Daniel and Castagn{\'e}, Roman and Luccioni, Alexandra Sasha and Yvon, Fran{\c{c}}ois and Gall{\'e}, Matthias and others},
  journal={arXiv preprint arXiv:2211.05100},
  year={2022}
}

@article{chowdhery2022palm,
  title={Palm: Scaling language modeling with pathways},
  author={Chowdhery, Aakanksha and Narang, Sharan and Devlin, Jacob and Bosma, Maarten and Mishra, Gaurav and Roberts, Adam and Barham, Paul and Chung, Hyung Won and Sutton, Charles and Gehrmann, Sebastian and others},
  journal={arXiv preprint arXiv:2204.02311},
  year={2022}
}

@misc{radford2018improving,
  title={Improving language understanding by generative pre-training},
  author={Radford, Alec and Narasimhan, Karthik and Salimans, Tim and Sutskever, Ilya and others},
  year={2018},
  publisher={OpenAI}
}

@article{radford2019language,
  title={Language models are unsupervised multitask learners},
  author={Radford, Alec and Wu, Jeffrey and Child, Rewon and Luan, David and Amodei, Dario and Sutskever, Ilya and others},
  journal={OpenAI blog},
  year={2019}
}

@article{gao2020pile,
  title={The pile: An 800gb dataset of diverse text for language modeling},
  author={Gao, Leo and Biderman, Stella and Black, Sid and Golding, Laurence and Hoppe, Travis and Foster, Charles and Phang, Jason and He, Horace and Thite, Anish and Nabeshima, Noa and others},
  journal={arXiv preprint arXiv:2101.00027},
  year={2020}
}

@misc{gpt-j,
  author = {Wang, Ben and Komatsuzaki, Aran},
  title = {{GPT-J-6B: A 6 Billion Parameter Autoregressive Language Model}},
  year = {2021},
  month = May
}

@inproceedings{bender2021dangers,
  title={On the Dangers of Stochastic Parrots: Can Language Models Be Too Big?},
  author={Bender, Emily M and Gebru, Timnit and McMillan-Major, Angelina and Shmitchell, Shmargaret},
  booktitle={Proceedings of the 2021 ACM Conference on Fairness, Accountability, and Transparency},
  pages={610--623},
  year={2021}
}

@inproceedings{carlini2021extracting,
  title={Extracting training data from large language models},
  author={Carlini, Nicholas and Tramer, Florian and Wallace, Eric and Jagielski, Matthew and Herbert-Voss, Ariel and Lee, Katherine and Roberts, Adam and Brown, Tom and Song, Dawn and Erlingsson, Ulfar and others},
  booktitle={30th USENIX Security Symposium (USENIX Security 21)},
  pages={2633--2650},
  year={2021}
}

@inproceedings{lin2021truthfulqa,
  author    = {Stephanie Lin and
               Jacob Hilton and
               Owain Evans},
  title     = {TruthfulQA: Measuring How Models Mimic Human Falsehoods},
  booktitle = {Proceedings of the 60th Annual Meeting of the Association for Computational
               Linguistics, {ACL}},
  pages     = {3214--3252},
  publisher = {Association for Computational Linguistics},
  year      = {2022},
  url       = {https://doi.org/10.18653/v1/2022.acl-long.229},
  doi       = {10.18653/v1/2022.acl-long.229},
  bibsource = {dblp computer science bibliography, https://dblp.org}
}

@article{taylor2022galactica,
  title={Galactica: A Large Language Model for Science},
  author={Taylor, Ross and Kardas, Marcin and Cucurull, Guillem and Scialom, Thomas and Hartshorn, Anthony and Saravia, Elvis and Poulton, Andrew and Kerkez, Viktor and Stojnic, Robert},
  journal={arXiv preprint arXiv:2211.09085},
  year={2022}
}

@online{chatgpt,
  title={ChatGPT: Optimizing Language Models for Dialogue},
  author={OpenAI},
  year={2022},
  url = {https://openai.com/blog/chatgpt/},
  urldate = {2022-12-28}
}

@book{beauchamp_principles_2019,
	address = {New York},
	edition = {Eighth edition},
	title = {Principles of biomedical ethics},
	isbn = {978-0-19-064087-3 978-0-19-008552-0},
	publisher = {Oxford University Press},
	author = {Beauchamp, Tom L. and Childress, James F.},
	year = {2019},
}

@article{hagendorff_ethics_2020,
	title = {The {Ethics} of {AI} {Ethics}: {An} {Evaluation} of {Guidelines}},
	volume = {30},
	issn = {0924-6495, 1572-8641},
	shorttitle = {The {Ethics} of {AI} {Ethics}},
	url = {http://link.springer.com/10.1007/s11023-020-09517-8},
	doi = {10.1007/s11023-020-09517-8},
	number = {1},
	urldate = {2022-12-23},
	journal = {Minds and Machines},
	author = {Hagendorff, Thilo},
	month = mar,
	year = {2020},
	pages = {99--120},
}

@inproceedings{DBLP:conf/acl/ZhouNGDGZG21,
  author    = {Chunting Zhou and
               Graham Neubig and
               Jiatao Gu and
               Mona T. Diab and
               Francisco Guzm{\'{a}}n and
               Luke Zettlemoyer and
               Marjan Ghazvininejad},
  title     = {Detecting Hallucinated Content in Conditional Neural Sequence Generation},
  booktitle = {Findings of the Association for Computational Linguistics: {ACL/IJCNLP}},
  pages     = {1393--1404},
  publisher = {Association for Computational Linguistics},
  year      = {2021},
  url       = {https://doi.org/10.18653/v1/2021.findings-acl.120},
  doi       = {10.18653/v1/2021.findings-acl.120},
  bibsource = {dblp computer science bibliography, https://dblp.org}
}

@article{oh2016porter,
  title={PORTER: a prototype system for patient-oriented radiology reporting},
  author={Oh, Seong Cheol and Cook, Tessa S and Kahn, Charles E},
  journal={Journal of digital imaging},
  volume={29},
  number={4},
  pages={450--454},
  year={2016},
  publisher={Springer}
}

@online{nyt,
  title={The Brilliance and Weirdness of ChatGPT},
  author={The New York Times},
  year={2022},
  url = {https://www.nytimes.com/2022/12/05/technology/chatgpt-ai-twitter.html},
  urldate = {2022-12-28}
}

@online{wapo,
  title={Stumbling with their words, some people let AI do the talking },
  author={The Washington Post},
  year={2022},
  url = {https://www.washingtonpost.com/technology/2022/12/10/chatgpt-ai-helps-written-communication/},
  urldate = {2022-12-28}
}

@online{bbc,
  title={ChatGPT: New AI chatbot has everyone talking to it},
  author={BBC},
  year={2022},
  url = {https://www.bbc.com/news/technology-63861322},
  urldate = {2022-12-28}
}

@online{guardian,
  title={What is AI chatbot phenomenon ChatGPT and could it replace humans?},
  author={The Guardian},
  year={2022},
  url = {https://www.theguardian.com/technology/2022/dec/05/what-is-ai-chatbot-phenomenon-chatgpt-and-could-it-replace-humans},
  urldate = {2022-12-28}
}


\appendix
\section{Appendix}
\label{appendix}
In Appendix \ref{appendix}, we summarize the original reports and the simplified reports generated by ChatGPT. The incorrect text passages (question 1a) identified by participants were highlighted in the simplified reports.  

\subsection{Report \#1: Knee MRI}
\addtocounter{subsubsection}{-1}

\subsubsection{Original Report}
\label{sec:appendix_original_report1}
\begin{tcolorbox}[]
Radiography is used for correlation.

\vspace{\baselineskip}
Extended MRI in different sequencing techniques and slice planes, application of additional, special sequencing techniques, including spectral and/or inversion-based fat suppression.

\vspace{\baselineskip}
Medial compartment:\\
Minor cartilage damage at dorsal femoral condyle grade 2. Minor edema subchondrally, medially at medial femoral condyle DD postcontusion. Meniscus intact. Higher grade partial rupture of superficial medial ligament portions. Specifically meniscofemoral somewhat frayed deep medial ligament portions. Decent distal muscle fiber tear of the semimembranosus muscle.

\vspace{\baselineskip}
Lateral compartment:\\
Extensive bone marrow edema at the lateral tibial plateau. No definite cartilage damage. Oblique tear at the posterior horn outer meniscus, tibial femoral side (3/8). Suspected damage to adjacent meniscopopliteal fascicles. Continuous lateral collateral ligament. If signal elevation is present, rupture of ALL suspected. Tractus iliotibialis continuous. Popliteus muscle intact.

\vspace{\baselineskip}
Patellofemoral joint/extensor:\\
Minor tendinopathy of quadriceps tendon, tendinopathy of patellar tendon at tibial tuberosity. Patella centered. M PFL thinned out.

\vspace{\baselineskip}
Intercondylar:
Anterior cruciate ligament cannot be continuously delineated, most likely complete rupture. Posterior cruciate ligament intact.

Conclusion:
\begin{enumerate}
\item[-] Complete rupture of the anterior cruciate ligament.
\item[-] Higher grade rupture of the medial ligament, predominantly of the superficial portions.
\item[-] Infraction at the lateral tibial plateau without step formation.
\item[-] Tear formation at the outer meniscus posterior horn. Ruptured anterolateral ligament.
\item[-] Muscle fiber tear of semimembranosus muscle.
Minimal cartilage damage to the medial femoral condyle.
\end{enumerate}
\end{tcolorbox}

\clearpage

\subsubsection{Simplified Report Sample \#1}
\label{app:rep1.1}

\begin{tcolorbox}[]
Radiography is a way to take pictures of the inside of your body using special machines.
\vspace{\baselineskip}

An MRI is a special kind of picture that shows more detail inside your body. It can be taken in different ways to get different types of pictures.
\vspace{\baselineskip}

The medial compartment is the inside part of your leg. There is some damage to the cartilage, which is a hard, smooth material that helps your joints move easily. There is also some swelling and bruising on the inside of your knee. The meniscus, which is a c-shaped piece of cartilage in your knee, is not damaged. However, there is some damage to the ligaments, which are strong bands of tissue that help hold your bones together. There is also a small tear in a muscle called the semimembranosus muscle.
\vspace{\baselineskip}

The lateral compartment is the outside part of your leg. There is a lot of swelling in the bone in the top part of your shin bone. There is no damage to the cartilage in this area. There is a tear in the meniscus, which is a c-shaped piece of cartilage in your knee, on the outside of your knee. There may also be some damage to the ligaments and muscles around your knee.
\vspace{\baselineskip}

The patellofemoral joint is the joint between your knee cap (patella) and your thigh bone (femur). There is some wear and tear on the tendons that attach your thigh muscles to your knee cap and shin bone. Your knee cap is in the right position. There is some thinning of a ligament called the M PFL.
\vspace{\baselineskip}

The intercondylar joint is the joint between the two rounded parts (condyles) on the top of your shin bone. The anterior cruciate ligament, which helps keep your shin bone in place, is most likely completely torn. The posterior cruciate ligament, which is on the back of your shin bone, is not damaged.
\vspace{\baselineskip}

In conclusion, you have a complete tear of the anterior cruciate ligament, a higher grade tear of the medial ligament, a tear in the meniscus, and a muscle fiber tear in the semimembranosus muscle. There is also some swelling in the bone on the outside of your shin and some wear and tear on the tendons around your knee cap. There is only minor damage to the cartilage on the inside of your thigh bone.
\end{tcolorbox}

\clearpage

\subsubsection{Simplified Report Sample \#2}
\label{app:rep1.2}

\begin{tcolorbox}[]
Radiography is a way of taking pictures of the inside of your body using special machines that use x-rays.
\vspace{\baselineskip}

An MRI is a way of taking pictures of the inside of your body using magnets and computers. It can be used to look at different parts of your body in different ways. Sometimes, it can be used to make the pictures clearer or to highlight certain parts of your body.
\vspace{\baselineskip}

The medial compartment is the inside part of your leg. There is some damage to the cartilage on the top of your shinbone and some swelling on the inside of your knee. The meniscus, which is a pad of cartilage in your knee, is okay. There is some damage to the ligaments on the inside of your knee, but they are not completely torn. There is also a small tear in one of the muscles in the back of your leg.
\vspace{\baselineskip}

The lateral compartment is the outside part of your leg. There is some swelling in the bone in the top of your shinbone on the outside of your knee. There is no damage to the cartilage. There is a small tear in one of the pads of cartilage in your knee. There might be some damage to the ligaments and muscles around your knee, but it is not certain. The ligaments on the outside of your knee are okay. The muscle on the back of your leg is okay.
\vspace{\baselineskip}

The patellofemoral joint is the joint between your knee cap and your thigh bone. There is some wear and tear on the tendons that attach your leg muscles to your bones. There is also some wear and tear on the tendon that attaches your knee cap to your shinbone. Your knee cap is in the right place. The ligament that holds your knee cap in place is a little thin.
\vspace{\baselineskip}

The intercondylar joint is the joint between the two parts of your thigh bone. The ligament that holds your knee together in the front is most likely completely torn. The ligament that holds your knee together in the back is okay.
\vspace{\baselineskip}

Overall, there is a complete tear in the ligament in the front of your knee and some tears in the ligaments and muscles on the inside and outside of your knee. There is also some damage to the bone and cartilage in your leg.
\end{tcolorbox}

\clearpage

\subsubsection{Simplified Report Sample \#3}
\label{app:rep1.3}

\begin{tcolorbox}[]
Radiography is a way to take pictures of the inside of your body using special machines.
\vspace{\baselineskip}

The doctor used a special kind of MRI machine to take many pictures of your leg from different angles. They also used some extra techniques to make the pictures clearer.
\vspace{\baselineskip}

In the middle part of your leg, there is a little bit of damage to the cartilage (a tough, rubbery substance that helps your joints move smoothly) at the top of your shin bone. There is also some swelling and bruising around the joint at the top of your shin bone. The cartilage in the middle of your knee is not damaged. The meniscus (a cushion in your knee) is not damaged either. However, some of the ligaments (tough, fibrous bands of tissue that hold your joints together) on the inside of your knee are slightly torn or frayed. There is also a small tear in a muscle called the semimembranosus muscle.
\vspace{\baselineskip}

On the outside of your leg, there is a lot of swelling in the bone around the joint at the top of your shin bone. The cartilage in this area is not damaged. There is a tear in the meniscus (a cushion in your knee) on the outside of your knee. It is possible that some other tissues around the meniscus are also damaged. The ligament on the outside of your knee (called the lateral collateral ligament) is not torn. However, there may be a tear in another ligament called the anterolateral ligament. The muscle called the popliteus is not damaged.
\vspace{\baselineskip}

In the front of your knee, the tendons (tough, fibrous cords that connect muscles to bones) that help straighten your leg are a little bit sore. The patella (your kneecap) is in the right place. There is some thinning of a ligament called the M PFL.
\vspace{\baselineskip}

In the back of your knee, the ligament that helps keep your shin bone in place (called the anterior cruciate ligament) is most likely completely torn. The ligament that helps keep your thigh bone in place (called the posterior cruciate ligament) is not damaged.
\vspace{\baselineskip}

Overall, it looks like you have a complete tear in the anterior cruciate ligament and higher grade tears in the medial ligaments in your knee. There is also a tear in the meniscus on the outside of your knee and a possible tear in the anterolateral ligament. There is a small tear in the semimembranosus muscle and minimal cartilage damage in the medial femoral condyle.
\end{tcolorbox}

\clearpage

\subsubsection{Simplified Report Sample \#4}
\label{app:rep1.4}

\begin{tcolorbox}[]
Radiography is a type of medical test that uses special machines to take pictures of the inside of your body.
\vspace{\baselineskip}

An MRI is another type of medical test that uses strong magnets and radio waves to take detailed pictures of the inside of your body. Sometimes, doctors will use different techniques or change the way the MRI machine takes pictures to get a better look at certain areas. They might also use special techniques to help them see certain tissues more clearly.
\vspace{\baselineskip}

The medical report is talking about a person's knee. The knee has two compartments, the medial compartment (the inner part) and the lateral compartment (the outer part).
\vspace{\baselineskip}

In the medial compartment, there is some minor damage to the cartilage (a smooth, rubbery material that covers the bones and helps them move smoothly against each other). There is also some swelling and minor damage to the bone in this area. The meniscus (a c-shaped piece of cartilage that acts like a shock absorber) is okay, but there is some damage to the ligaments (tough bands of tissue that hold the bones together). There is also a small tear in the semimembranosus muscle (a muscle in the leg).
\vspace{\baselineskip}

In the lateral compartment, there is some swelling in the bone and no definite cartilage damage. There is also a tear in the meniscus and possible damage to other tissues. The lateral collateral ligament (another ligament in the knee) is okay. There is a possibility of damage to the anterolateral ligament (another ligament in the knee), and the popliteus muscle (another muscle in the leg) is okay.

\vspace{\baselineskip}
In the patellofemoral joint (the joint between the patella, or kneecap, and the femur, or thigh bone) and the quadriceps tendon (a tendon that connects the quadriceps muscle to the patella), there is some minor wear \hlc{and tear.} The patella (kneecap) is in the correct position. The medial patellofemoral ligament (a ligament in the knee) is thinned out.
\vspace{\baselineskip}

In the intercondylar area (\hlc{the area between the two bones that make up the lower part of} the leg), the anterior cruciate ligament (a ligament in the knee) is most likely completely ruptured (broken). The posterior cruciate ligament (another ligament in the knee) is okay.
\vspace{\baselineskip}

The conclusion of the medical report is that the person has a complete rupture of the anterior cruciate ligament, higher grade rupture of the medial ligament (mostly the superficial portions), a small fracture in the lateral tibial plateau without a step formation, a tear in the outer meniscus posterior horn, and a muscle fiber tear of the semimembranosus muscle. There is also minimal cartilage damage to the medial femoral condyle.
\end{tcolorbox}

\clearpage

\subsubsection{Simplified Report Sample \#5}
\label{app:rep1.5}

\begin{tcolorbox}[]
Radiography is a way doctors can take pictures of the inside of your body using special machines.
\vspace{\baselineskip}

The doctors used a special machine called an MRI to take detailed pictures of your legs. They looked at different parts of your legs and used different techniques to get the best pictures.
\vspace{\baselineskip}

\hlc{In the middle part of your leg} (called the medial compartment), there was a small amount of damage to some cartilage on your knee and a small amount of swelling. The meniscus, which is a small, crescent-shaped piece of cartilage in your knee, was not damaged. However, some of the ligaments that help hold your knee together were partially torn. There was also a small tear in a muscle called the semimembranosus muscle.
\vspace{\baselineskip}

In the outer part of your leg (called the lateral compartment), there was a lot of swelling in the bone in your lower leg and no damage to the cartilage. There was a tear in the meniscus in your knee on the outer side. There might also be some damage to some other ligaments and tissues near the meniscus. The lateral collateral ligament was not damaged, but there might be a problem with the anterolateral ligament. The muscle called the popliteus was not damaged.
\vspace{\baselineskip}

In the front of your knee (called the patellofemoral joint), there was a small amount of wear and tear on the tendons that connect your thigh muscle to your knee and shin bone. The patella, or kneecap, was in the right place. There was also some thinning of the ligament called the M PFL.
\vspace{\baselineskip}

In the area between the bones of your knee (called the intercondylar area), the anterior cruciate ligament was most likely completely torn. The posterior cruciate ligament was not damaged.
\vspace{\baselineskip}

Overall, the main problems were a complete tear in the anterior cruciate ligament, a higher grade tear in the medial ligament, a tear in the meniscus in the outer part of the knee, a small tear in a muscle in the middle part of the leg, and some swelling in the bone in the outer part of the leg. There was only a small amount of damage to the cartilage in the \hlc{middle part of the leg.}
\end{tcolorbox}

\clearpage

\subsubsection{Simplified Report Sample \#6}
\label{app:rep1.6}

\begin{tcolorbox}[]
Radiography is a test that uses X-rays to take pictures of the inside of the body. The doctors used this test to get more information about the patient's injuries.
\vspace{\baselineskip}

The MRI is a test that uses strong magnets and radio waves to take detailed pictures of the inside of the body. The doctors used different techniques and took pictures from different angles to get a better look at the patient's injuries.
\vspace{\baselineskip}

The patient had some injuries to their knee. There was some damage to the cartilage (a hard, smooth surface that covers the ends of bones) on the top part of the shinbone. There was also some swelling and injury to the cartilage on the inside of the knee. The meniscus (a cushion in the knee) was not damaged. However, some of the ligaments (tough bands of tissue that hold bones together) on the inside of the knee were partly torn. The patient also had a small tear in a muscle called the semimembranosus muscle.
\vspace{\baselineskip}

On the outside of the knee, there was swelling in the bone and some damage to the meniscus. There might also be damage to other tissues in the knee. The ligaments on the outside of the knee were not torn. The popliteus muscle (a muscle in the back of the knee) was not damaged.
\vspace{\baselineskip}

The patient also had some injuries to the tendons (tough cords of tissue that connect muscles to bones) in their knee. The quadriceps tendon (a tendon in the front of the thigh) and the patellar tendon (a tendon in the front of the shinbone) were slightly damaged. The patella (kneecap) was in the correct position. The ligament that helps hold the patella in place was slightly damaged.
\vspace{\baselineskip}

In the middle part of the knee, the anterior cruciate ligament (a ligament that helps keep the knee stable) was completely torn. The posterior cruciate ligament (another ligament that helps keep the knee stable) was not damaged.
\vspace{\baselineskip}

In summary, the patient had a complete tear of the anterior cruciate ligament and partial tears of the ligaments on the inside of the knee. There was also some damage to the cartilage on the inside of the knee, swelling in the bone on the outside of the knee, and a tear in the meniscus on the outside of the knee. The patient also had some slight damage to the tendons in the front of the leg and the patella.
\end{tcolorbox}

\clearpage

\subsubsection{Simplified Report Sample \#7}
\label{app:rep1.7}

\begin{tcolorbox}[]
Okay, let's break this down into simpler language.
\vspace{\baselineskip}

Radiography is a type of medical test that uses special machines to take pictures of the inside of your body. This can help doctors understand what might be wrong and how to help you.
\vspace{\baselineskip}

An MRI is another type of medical test that also takes pictures of the inside of your body. It uses magnets and radio waves instead of x-rays to get these pictures. Sometimes, doctors need to do a special type of MRI where they take even more detailed pictures from different angles, or they use extra techniques to get a clearer image.
\vspace{\baselineskip}

Your knee has different parts, or compartments, that can be injured. The medial compartment is the inside part of your knee, and the lateral compartment is the outside part.
\vspace{\baselineskip}

In the medial compartment, there is some minor damage to the cartilage on the top of your thigh bone. Cartilage is a tough, slippery material that helps your bones move smoothly against each other. There is also some swelling and a little bit of injury to the cartilage on the inside part of your knee. The meniscus, which is a C-shaped piece of cartilage in your knee, is not damaged. However, there is some damage to the ligaments, which are strong bands of tissue that hold your bones together and help your knee move. There is also a small tear in one of the muscles in your thigh.
\vspace{\baselineskip}

In the lateral compartment, there is a lot of swelling and injury \hlc{in the bone inside your lower leg bone.} There is no damage to the cartilage in this area. There is a tear in the meniscus, which is the C-shaped piece of cartilage in your knee, on the outside part of your knee. There might also be some damage to other tissues in this area. The ligament on the outside of your knee is not damaged. There is also no damage \hlc{to the muscle in the back of your knee}.
\vspace{\baselineskip}

In the patellofemoral joint, which is the joint where your kneecap (patella) and thigh bone (femur) meet, there is some minor damage to the tendons, which are strong cords of tissue that attach muscles to bones. The kneecap is in the right place. There is also some thinning of the ligament that helps hold the kneecap in place.
\vspace{\baselineskip}

In the intercondylar area, which is the area between the two bumpy parts (condyles) at the \hlc{top of your thigh bone,} the anterior cruciate ligament (ACL) is completely torn. The ACL is a very important ligament that helps keep your knee stable. The posterior cruciate ligament (PCL), which is another important ligament in your knee, is not damaged.
\vspace{\baselineskip}

Overall, this report is saying that you have a complete tear of the ACL and some other injuries to the ligaments, bones, and cartilage in your knee. You also have a small tear in a muscle in your thigh and some minor damage to the tendons around your kneecap. There is minimal damage to the cartilage on the inside part of your knee.
\end{tcolorbox}

\clearpage

\subsubsection{Simplified Report Sample \#8}
\label{app:rep1.8}

\begin{tcolorbox}[]
Radiography is a way to take pictures of the inside of your body using special equipment. MRI stands for magnetic resonance imaging. It's a way to get even more detailed pictures of the inside of your body by using magnets and radio waves. Different techniques and slice planes are just different ways to take the pictures.
\vspace{\baselineskip}

The medial compartment is the inner part of your knee. There is some damage to the cartilage, which is the smooth layer on top of the bone that helps your knee move smoothly. There is also some swelling and bruising on the inner part of the bone. The meniscus, which is a cushion in your knee, is okay. But there is some damage to the ligaments, which are strong bands of tissue that hold your knee together. There is also a small tear in one of the muscles that helps you bend your leg.
\vspace{\baselineskip}

The lateral compartment is the outer part of your knee. There is a lot of swelling in the bone and some damage to the meniscus, which is the cushion in your knee. There is also possible damage to other tissues in your knee. The lateral collateral ligament, which is another strong band of tissue that helps keep your knee stable, is okay. The iliotibial band and the popliteus muscle, which are other tissues that help your leg move, are also okay.
\vspace{\baselineskip}

The patellofemoral joint is the joint where your kneecap (patella) moves against your thigh bone (femur). There is some minor irritation of the tendons, which are strong cords of tissue that connect your muscles to your bones. The kneecap is in the right position. The medial patellofemoral ligament, which is a strong band of tissue that helps keep your kneecap in place, is thinned out.
\vspace{\baselineskip}

The intercondylar area is the space between the bones (condyles) in your knee. The anterior cruciate ligament, which helps keep your knee stable when you move, is most likely completely torn. The posterior cruciate ligament, which is another ligament that helps keep your knee stable, is okay.
\vspace{\baselineskip}

Overall, there is a complete tear of the anterior cruciate ligament and some other damage to the ligaments and muscles in your knee. There is also some swelling in the bone and minor damage to the cartilage.
\end{tcolorbox}

\clearpage

\subsubsection{Simplified Report Sample \#9}
\label{app:rep1.9}

\begin{tcolorbox}[]
Radiography is a medical test that takes pictures of the inside of your body. It helps doctors understand what is happening inside your body.
\vspace{\baselineskip}

An MRI is another medical test that takes pictures of the inside of your body. It uses strong magnets and radio waves to create detailed pictures of your bones and tissues. The MRI in this report used different ways of taking pictures and looked at different parts of the body.
\vspace{\baselineskip}

The report mentions several parts of the body that were examined, including the medial compartment, which is the \hlc{inside} part of the knee, and the lateral compartment, which is the \hlc{outside} part of the knee. The report also mentions the patellofemoral joint, which is the joint between the kneecap and the thigh bone, and the intercondylar, which is the area between the two bones that make up the lower part of the leg.
\vspace{\baselineskip}

The report says that there is some damage to the cartilage, which is a \hlc{hard, smooth} substance that helps protect the bones, in the medial compartment. There is also swelling and some damage to the ligaments, which are strong bands of tissue that hold the bones together, in the medial compartment. The meniscus, which is a crescent-shaped piece of cartilage in the knee, is not damaged. However, there is a tear in the muscle called the semimembranosus muscle.
\vspace{\baselineskip}

In the lateral compartment, there is swelling in the bone and a tear in the meniscus. There may also be damage to the ligaments and muscle called the iliotibialis. The popliteus muscle is not damaged.
\vspace{\baselineskip}

In the patellofemoral joint and extensor, there is some wear and tear on the tendons, which are strong cords of tissue that connect muscles to bones, and the patella, which is the kneecap, is in the correct position.
\vspace{\baselineskip}

In the intercondylar, the anterior cruciate ligament, which is a ligament in the knee, is most likely completely torn. The posterior cruciate ligament, which is another ligament in the knee, is not damaged.
\vspace{\baselineskip}

Overall, the report says that there is a complete tear in the anterior cruciate ligament and some damage to other ligaments and muscles in the knee. There is also some swelling in the bone and a tear in the meniscus. However, there is minimal damage to the cartilage in the medial compartment.
\end{tcolorbox}

\clearpage

\subsubsection{Simplified Report Sample \#10}
\label{app:rep1.10}

\begin{tcolorbox}[]
Radiography is a special kind of test that uses x-rays to create pictures of the inside of your body. The doctors used this test to help them understand what is happening in your body.
\vspace{\baselineskip}

The MRI is another special test that uses magnets and computers to create detailed pictures of the inside of your body. The doctors used different techniques and planes to get a good look at different parts of your body.
\vspace{\baselineskip}

In the medial compartment (the inside part) of your leg, there is some minor damage to the cartilage (the smooth, white material that covers the ends of bones) on the top part of your thigh bone. There is also some swelling and minor damage to the cartilage on the inside part of the bone at the bottom of your thigh. The meniscus (the c-shaped piece of cartilage in your knee) is okay. However, there is some damage to the ligaments (the strong, elastic bands of tissue that hold bones together) on the inside of your leg. There is also some damage to the muscle (the tissues in your body that help you move) on the back part of your thigh.
\vspace{\baselineskip}

In the lateral compartment (the outside part) of your leg, there is some swelling in the bone marrow (the spongy tissue inside your bones that helps make blood cells) on the outer part of your shin bone. There is no damage to the cartilage in this area. However, there is a small tear in the meniscus (the c-shaped piece of cartilage in your knee) on the outer part of your leg. There might also be some damage to the tissues around the meniscus. The ligament (the strong, elastic bands of tissue that hold bones together) on the outside of your leg is okay. The muscle (the tissues in your body that help you move) on the back of your leg is also okay.
\vspace{\baselineskip}

In the patellofemoral joint (the joint where your knee cap meets your thigh bone) and the extensor muscles (the muscles that help you straighten your leg), there is some minor damage to the tendons (the tough, white cords that connect your muscles to your bones) in your quadriceps (the group of muscles at the front of your thigh) and your knee cap. Your knee cap is in the right position. The ligament (the strong, elastic bands of tissue that hold bones together) on the inside of your leg is thinned out.
\vspace{\baselineskip}

In the intercondylar (the space between the two parts of your thigh bone) area, the anterior cruciate ligament (ACL, one of the main ligaments in your knee) is probably completely torn. The posterior cruciate ligament (PCL, another main ligament in your knee) is okay.
\vspace{\baselineskip}

In conclusion, you have a complete tear of the ACL ligament and some higher grade tears of the ligaments on the inside of your leg. There is also some damage to the bone on the outer part of your shin and a tear in the meniscus on the outer part of your leg. You have a small tear in the muscle on the back of your thigh and some minor damage to the cartilage on the inside part of the bone at the bottom of your thigh.
\end{tcolorbox}

\clearpage

\subsubsection{Simplified Report Sample \#11}
\label{app:rep1.11}

\begin{tcolorbox}[]
This medical report is talking about a person's knee. The knee is made up of different parts, including the thigh bone, shin bone, and kneecap. The report says that there are some problems with the knee.
\vspace{\baselineskip}

One problem is that there is a tear in the ligament, which is a strong band of tissue that connects bones. The ligament on the inside of the knee has a higher grade tear, which means it is more serious. There is also a tear in a muscle called the semimembranosus muscle.
\vspace{\baselineskip}

Another problem is that the knee has some swelling, which is called edema. There is also some damage to the cartilage, which is a smooth, white tissue that covers the ends of the bones and helps them move smoothly.
\vspace{\baselineskip}

There is also a problem with the knee joint on the side of the knee. There is swelling in the bone and a tear in the meniscus, which is a C-shaped piece of cartilage that helps cushion the joint. There may also be a problem with the ligament on the side of the knee.
\vspace{\baselineskip}

Finally, the report says that there may be a problem with the ligaments that connect the thigh bone to the shin bone. One of these ligaments, called the anterior cruciate ligament, may be completely torn. However, the other ligament, called the posterior cruciate ligament, is still okay.
\vspace{\baselineskip}

Overall, this medical report is saying that the person has some injuries to their knee that will need to be treated.
\end{tcolorbox}

\clearpage

\subsubsection{Simplified Report Sample \#12}
\label{app:rep1.12}

\begin{tcolorbox}[]
Radiography is a type of medical test that helps doctors see inside your body using special machines. They use it to help them understand what might be causing a problem or injury.
\vspace{\baselineskip}

An MRI is another type of medical test that uses powerful magnets and radio waves to take pictures of the inside of your body. There are different ways the doctors can use the MRI machine to get different types of pictures. They might also use other special techniques to get even clearer pictures.
\vspace{\baselineskip}

The medial compartment is the inside part of your leg. There is some damage to a small piece of cartilage on the top part of the bone in your leg. There is also some swelling and a little bit of injury to the bone under the cartilage. The meniscus, which is a small, C-shaped piece of cartilage in your knee, is not injured. But some of the ligaments that help hold your knee together are a little bit damaged or frayed. There is also a small tear in one of the muscles in your leg.
\vspace{\baselineskip}

The lateral compartment is the outside part of your leg. There is some swelling in the bone in the top part of your leg. There is no damage to the cartilage in your knee. There is a small tear in a part of the meniscus that is on the outside of your leg. There might also be some damage to other small ligaments near the meniscus. The ligament that helps hold your leg and knee together is not damaged. The muscle that helps you bend your leg is also not damaged.
\vspace{\baselineskip}

The patellofemoral joint is the joint between your kneecap and your leg bone. There is some minor injury to the tendons in your leg that help you straighten your leg. There is also some injury to the tendon that goes from your kneecap to your leg bone. Your kneecap is in the right position. There is some thinning of a ligament that helps hold your kneecap in place.
\vspace{\baselineskip}

The intercondylar is the area between the two bones in your leg. The ligament that helps hold your leg bones together in the front is most likely completely torn. The ligament that helps hold your leg bones together in the back is not damaged.
\vspace{\baselineskip}

The conclusion is that you have a complete tear in the front ligament in your knee, a higher grade tear in some of the ligaments on the inside of your leg, some swelling in the bone on the outside of your leg without any step formation, a tear in the meniscus on the outside of your leg, and a small tear in one of the muscles in your leg. There is only minimal damage to the cartilage on the inside of your leg bone.
\end{tcolorbox}

\clearpage

\subsubsection{Simplified Report Sample \#13}
\label{app:rep1.13}

\begin{tcolorbox}[]
Radiography is a way to take pictures inside the body using special machines. The pictures help doctors see what is going on inside the body.
\vspace{\baselineskip}

MRI is another way to take pictures inside the body. It uses strong magnets and radio waves to make the pictures. Sometimes, doctors need to use different techniques or planes to get the best pictures.
\vspace{\baselineskip}

The medial compartment is the part of the knee on the inside. In this case, there is some damage to the cartilage, which is the smooth white material that helps the bones move smoothly. There is also some swelling and bruising on the bone under the cartilage. The meniscus, which is a c-shaped piece of cartilage in the knee, is okay. But some of the ligaments, which are strong bands of tissue that help hold the bones together, are a little bit damaged. There is also a small tear in a muscle called the semimembranosus muscle.
\vspace{\baselineskip}

The lateral compartment is the part of the knee on the outside. In this case, there is a lot of swelling in the bone under the cartilage \hlc{on the outside of the tibia,} which is one of the bones in the leg. There is no damage to the cartilage, but there is a tear in the meniscus, which is a c-shaped piece of cartilage in the knee. There might also be damage to some of the ligaments and muscle fibers in this area. The ligaments on the outside of the knee, called the lateral collateral ligament, and the muscle called the popliteus muscle, are both okay.
\vspace{\baselineskip}

The patellofemoral joint is the joint where the kneecap (patella) and thighbone (femur) meet. In this case, there is some minor damage to the tendons that attach the quadriceps muscles (which straighten the leg) and the patellar tendon (which extends the leg) to the bones. The kneecap is in the right position. The medial patellofemoral ligament (which helps hold the kneecap in place) is a little bit thin.
\vspace{\baselineskip}

The intercondylar joint is the joint between the two bones (tibia and femur) in the lower leg. In this case, the anterior cruciate ligament (which helps hold the bones together and keeps the knee stable) is most likely completely torn. The posterior cruciate ligament (which also helps hold the bones together and keeps the knee stable) is still intact.
\vspace{\baselineskip}

In conclusion, there is a complete tear of the anterior cruciate ligament and a higher grade tear of the medial ligament in the knee. There is also a tear in the meniscus and some damage to the lateral collateral ligament on the outside of the knee. There is a small tear in the semimembranosus muscle and minimal damage to the cartilage on the inside of the knee.
\end{tcolorbox}

\clearpage

\subsubsection{Simplified Report Sample \#14}
\label{app:rep1.14}

\begin{tcolorbox}[]
Radiography is a way to take pictures of the inside of your body using special machines. MRI stands for magnetic resonance imaging and is another way to take pictures of the inside of your body using magnets and radio waves. Different slices and techniques can be used to get different kinds of pictures.
\vspace{\baselineskip}

The medial compartment is the part of your leg that is closest to the middle of your body. There is some minor damage to the cartilage on the top of your thigh bone, and there is some swelling near the cartilage on the top of the bone in your lower leg. The meniscus, which is a piece of cartilage in your knee, is okay. There is also some damage to some of the ligaments on the inside of your knee. The ligaments are like strong ropes that help hold your bones together. There is also a small tear in a muscle called the semimembranosus muscle.
\vspace{\baselineskip}

The lateral compartment is the part of your leg that is farther from the middle of your body. There is some swelling in the bone in the top of your lower leg, but no damage to the cartilage. There is a tear in a piece of cartilage called the meniscus on the outside of your knee. There might also be some damage to some other ligaments and muscles near your knee. The ligaments and muscles on the outside of your knee are okay.
\vspace{\baselineskip}

The patellofemoral joint is the joint between your kneecap (patella) and your thigh bone (femur). There is some minor damage to the tendons, which are like strong cords that connect your muscles to your bones, in your knee. Your kneecap is in the right place. There is also some thinning of a ligament called the M PFL.
\vspace{\baselineskip}

The intercondylar area is the space between the two parts of your thigh bone that stick out at the top of your knee. The anterior cruciate ligament (ACL), which is a ligament that helps hold your thigh bone and lower leg bone together, is probably completely torn. The posterior cruciate ligament (PCL), which is another ligament that helps hold your thigh bone and lower leg bone together, is okay.
\vspace{\baselineskip}

In conclusion, you have a complete tear of the ACL and a higher grade tear of the medial ligament in your knee. There is also a tear in the meniscus and possibly some damage to other ligaments and muscles on the outside of your knee. There is a small tear in the semimembranosus muscle and minimal cartilage damage on the inside of your knee.
\end{tcolorbox}

\clearpage

\subsubsection{Simplified Report Sample \#15}
\label{app:rep1.15}

\begin{tcolorbox}[]
Sure, I can explain this medical report to a child using simple language.
\vspace{\baselineskip}

Radiography is a way to take pictures of the inside of your body, usually using x-rays.
\vspace{\baselineskip}

An MRI is a special machine that uses powerful magnets and radio waves to take detailed pictures of the inside of your body. It can show different parts of your body in different ways by using different techniques, like special slicing planes and special techniques to highlight or suppress certain types of tissue.
\vspace{\baselineskip}

The \hlc{medial compartment} is the part of \hlc{your body} that's on the inside, closer to the center. There's some minor damage to the cartilage (a special kind of tissue that helps protect bones) on the top of \hlc{your leg bone,} and some swelling on the inside of your \hlc{leg bone} near the top. The meniscus (a special kind of tissue that helps protect your knee joint) is okay. There's also some damage to the ligaments (tissues that hold bones together) on the inside of your knee, and a small tear in the muscle (tissue that helps you move) on the back of your thigh.
\vspace{\baselineskip}

The \hlc{lateral compartment is the part of your body that's on the outside}, farther from the center. There's some swelling in the bone tissue on the outside of your lower leg bone, but no damage to the cartilage. There's also a tear in the meniscus on the outside of your knee, and possibly some damage to nearby tissues. The ligament on the outside of your knee is okay, but there might be a tear in the ligament on the front of your thigh. The muscle on the back of your calf is okay.
\vspace{\baselineskip}

The patellofemoral joint is the joint where your kneecap (patella) and your thigh bone (femur) meet. There's some minor damage to the tendons (tissues that connect muscles to bones) on the front of your thigh and your kneecap, but the kneecap is in the right place. The ligament on the inside of your knee is a little thin.
\vspace{\baselineskip}

The intercondylar joint is the joint between the two parts of your thigh bone (femur). The anterior cruciate ligament (ACL) is probably completely torn. The posterior cruciate ligament (PCL) is okay.
\vspace{\baselineskip}

So, the main problems are: a complete tear of the ACL, a higher grade tear of the medial ligament, a tear in the meniscus on the outside of the knee, a small tear in the muscle on the back of the thigh, and some minor damage to the cartilage on the inside of the leg bone.
\end{tcolorbox}

\clearpage

\subsection{Report \#2: Head MRI}
\addtocounter{subsubsection}{-1}

\subsubsection{Original Report}
\label{sec:appendix_original_report2}

\begin{tcolorbox}[]
Several previous examinations are available for comparison.
\vspace{\baselineskip}

The contrast-enhancing mass on the right parietooccipital side shows a progression in size in the short-term follow-up (currently max. 22 mm, in the previous examination max. 12 mm), with significantly increasing perifocal edema. The resection area on the right frontal side remains essentially unchanged, and there is still no definite evidence of a new contrast-enhancing mass. Otherwise, no evidence of recent or intermediate ischemia. No evidence of intracranial hemorrhage. No signs of cerebrospinal fluid circulation disturbance.
\vspace{\baselineskip}

Conclusion:\\
- In the short-term follow-up, significantly size-progressive contrast-enhancing space-occupying lesion right parietooccipital with significantly increasing perifocal edema, DD distant GBM manifestation.
No evidence of a contrast-enhancing mass in the right frontal resection area.

\end{tcolorbox}

\clearpage

\subsubsection{Simplified Report Sample \#1}
\label{app:rep2.1}

\begin{tcolorbox}[]
This medical report is talking about a person's brain. There have been some previous exams that doctors can compare to the current one.
\vspace{\baselineskip}

There is a mass in the right side of the brain that has gotten bigger since the last exam. It is also causing swelling around it. \hlc{There is no new mass in the right side of the brain,} but there is no evidence of any problems with the blood vessels or bleeding in the brain. There are also no problems with the fluid that surrounds the brain.
\vspace{\baselineskip}

The doctors think that the mass in the right side of the brain is likely a type of brain cancer called glioblastoma.
\end{tcolorbox}

\clearpage

\subsubsection{Simplified Report Sample \#2}
\label{app:rep2.2}

\begin{tcolorbox}[]
This medical report is talking about a person's brain. There have been previous times when doctors have looked at the brain and taken pictures of it. The report says that there is a mass (a lump) on the right side of the back of the head that has grown bigger in size recently. It also says that there is swelling (edema) around the mass. The report also mentions that there was an area on the right side of the front of the head that the doctors removed something from before, and there is no new mass there now. The report also says that there is no evidence of bleeding in the brain or any problems with the fluid that surrounds the brain. The conclusion of the report says that the mass on the right side of the back of the head is likely a type of brain cancer called glioblastoma.
\end{tcolorbox}

\clearpage

\subsubsection{Simplified Report Sample \#3}
\label{app:rep2.3}

\begin{tcolorbox}[]
Okay, let's break down this medical report into simpler language.
\vspace{\baselineskip}

This report is about a person who has had some medical tests done before and has had more tests done recently. The tests show that there is a small, abnormal growth in the person's brain that has gotten bigger in the short-term follow-up (recent tests). This growth is called a "mass" and it is causing swelling around it. There was no change in another area of the brain that had been treated before, and there is no new growth in that area. The tests also showed that there is no bleeding in the brain and no problems with the fluid that surrounds the brain.
\vspace{\baselineskip}

The conclusion of the report is that the person has a \hlc{small,} abnormal growth in their brain that has gotten \hlc{bigger} recently and is causing swelling. This growth is called a "space-occupying lesion" and it is likely a type of brain cancer called glioblastoma (GBM). There is no evidence of a similar growth in the other area of the brain that was treated before.
\end{tcolorbox}

\clearpage

\subsubsection{Simplified Report Sample \#4}
\label{app:rep2.4}

\begin{tcolorbox}[]
This medical report is talking about a person's brain. The report says that there is a mass or growth in a certain area of the brain that has gotten bigger since the last time it was checked. There is also \hlc{swelling} around the mass. The report also says that there is no evidence of any other growths in a different area of the brain that was previously treated. The report also mentions that there is no evidence of any problems with the brain's blood flow or with the fluid that surrounds the brain. The report concludes that the mass is likely a type of brain cancer called a glioblastoma.
\end{tcolorbox}

\clearpage

\subsubsection{Simplified Report Sample \#5}
\label{app:rep2.5}

\begin{tcolorbox}[]
This medical report is talking about a person's brain. There are some pictures of the brain from different times, and the doctors are comparing them to see how the brain has changed.
\vspace{\baselineskip}

There is a problem in the right side of the back of the brain. It is called a "mass," and it is bigger now than it was before. It is also causing some swelling around it. On the right side of the front of the brain, there is a place where the doctors took out some tissue before, and it has not changed. There is no new problem in that area.
\vspace{\baselineskip}

Overall, the brain does not have any signs of not getting enough blood or oxygen, or bleeding. The fluid that surrounds the brain and spine is also normal.
\vspace{\baselineskip}

The doctors think that the problem in the right side of the back of the brain might be a type of tumor called a "glioblastoma." This is a serious type of tumor that can spread to other parts of the brain.
\vspace{\baselineskip}

The report concludes by saying that the problem in the right side of the back of the brain has gotten bigger and is causing swelling, and it might be a glioblastoma. There is no new problem in the right side of the front of the brain.
\end{tcolorbox}

\clearpage

\subsubsection{Simplified Report Sample \#6}
\label{app:rep2.6}

\begin{tcolorbox}[]
This medical report is about someone who has had a few medical exams in the past, and the doctors are comparing the results from these exams to see if anything has changed.
\vspace{\baselineskip}

The report says that there is a mass (a group of \hlc{abnormal} cells) on the right side of the back of the head that has gotten bigger since the last exam. The mass is called a "contrast-enhancing mass" because it shows up better on certain types of medical images when a special \hlc{dye} is used. The mass has grown from 12 millimeters (mm) to 22 mm, which is a big change in a short amount of time. There is also swelling (edema) around the mass.
\vspace{\baselineskip}

On the other side of the head, in the right frontal area, there was a surgery to remove a mass. The report says that this area looks the same as it did before and there is no evidence of a new mass forming.
\vspace{\baselineskip}

The report also mentions that there is no evidence of bleeding in the brain or any problems with the fluid that surrounds the brain and spine.
\vspace{\baselineskip}

The conclusion of the report is that the mass on the right side of the head \hlc{is} a type of cancer called a glioblastoma, and it has gotten bigger and caused more swelling in a short amount of time. There is no evidence of a new mass in the area where the surgery was done.
\end{tcolorbox}

\clearpage

\subsubsection{Simplified Report Sample \#7}
\label{app:rep2.7}

\begin{tcolorbox}[]
This medical report is talking about a person's brain. They have had some tests done before and are comparing them to a new test.
\vspace{\baselineskip}

There is a problem in the right side of the back of the \hlc{head.} It is a mass (a lump) that is growing. It is also causing swelling around it.
\vspace{\baselineskip}

There is no new problem in the right side of the front of the \hlc{head,} and there are no problems with the blood vessels or bleeding in the brain. There are also no problems with the fluid that surrounds the brain.
\vspace{\baselineskip}

The doctors think that the mass on the right side of the back of the head is a type of brain cancer called a GBM.
\end{tcolorbox}

\clearpage

\subsubsection{Simplified Report Sample \#8}
\label{app:rep2.8}

\begin{tcolorbox}[]
This medical report is talking about a mass (a lump) that was found in the brain. This mass is growing bigger and bigger, and it is causing swelling around it. The doctors have been watching it over time, and they can see that it has gotten bigger in the short term (recently). The mass is located on the right side of the back of the head. There is no other mass in the right side of the front part of the head. The doctors do not see any problems with blood flow or bleeding in the brain, and they do not see any problems with the fluid that surrounds the brain.
\vspace{\baselineskip}

The doctors \hlc{think that this mass might be a type of brain tumor called a glioblastoma (GBM).} It is called a "space-occupying lesion" because it takes up space in the brain and can cause problems. The doctors will continue to watch the mass and see if it gets bigger or if there are any other changes.
\end{tcolorbox}

\clearpage

\subsubsection{Simplified Report Sample \#9}
\label{app:rep2.9}

\begin{tcolorbox}[]
This medical report is about a person who has had some medical tests done on their brain. The report shows that there is a mass (a lump) on the right side of the brain that has gotten bigger since the last time it was checked. The mass is causing some swelling around it. There is no new mass on the right side of the head, and the person's brain \hlc{does not seem to be damaged or} bleeding. There is also no problem with the fluid that surrounds the brain. The doctors think that the mass might be a type of cancer called a glioblastoma.
\end{tcolorbox}

\clearpage

\subsubsection{Simplified Report Sample \#10}
\label{app:rep2.10}

\begin{tcolorbox}[]
This medical report is talking about a person who has had some medical tests done recently, and they also had some tests done before. The doctors are comparing the results of the new tests to the old ones to see if anything has changed.
\vspace{\baselineskip}

The report says that there is a mass (a lump) on the right side of the back of the \hlc{head} that has gotten bigger since the last time they were checked. The mass is called a "contrast-enhancing mass" because it shows up better on the medical tests when a special dye is used. The mass is also causing some swelling around it.
\vspace{\baselineskip}

The report also says that there is no new mass on the right side of the front of the head, and that there are no signs of any problems with blood flow to the brain or with bleeding in the brain. There are also no signs of any problems with the fluid that surrounds the brain.
\vspace{\baselineskip}

The doctors have concluded that the mass on the back of the \hlc{head} is likely a type of brain tumor called a glioblastoma, and it is getting bigger. There is no new mass on the \hlc{front of the head}.
\end{tcolorbox}

\clearpage

\subsubsection{Simplified Report Sample \#11}
\label{app:rep2.11}

\begin{tcolorbox}[]
 A medical report is a document that describes the results of a medical examination or test. It can be written in medical language, which is a special way of talking about the body and its functions.
\vspace{\baselineskip}

This medical report is talking about a person's brain. The brain is the part of the body that controls everything we do and think.
\vspace{\baselineskip}

The report says that there is a problem in the right side of the brain. It's a mass, which means it's a lump or bump. This mass is growing bigger and causing swelling around it. This is called edema.
\vspace{\baselineskip}

The report also mentions that there was a surgery done on the right side of the brain before. The report says that there is no new problem in that area.
\vspace{\baselineskip}

The report also says that there is no bleeding in the brain, and that the brain is getting enough blood and oxygen. It also says that the brain is not having any trouble sending messages through the body.
\vspace{\baselineskip}

The conclusion of the report is that there is a growing mass on the right side of the brain that is causing swelling, and it might be a type of cancer called a GBM. But there is no new problem in the area where the surgery was done.
\end{tcolorbox}

\clearpage

\subsubsection{Simplified Report Sample \#12}
\label{app:rep2.12}

\begin{tcolorbox}[]
Okay, let's break this down into simpler language:
\vspace{\baselineskip}

There are some past medical exams that we can compare to the current one. The current exam shows that there is a mass (a lump) on the right side of the back of the head that has grown bigger since the last exam (it's 22mm now, but it was 12mm before). There is also some swelling around the mass. There is no new mass on the right side of the front of the head, and there are no problems with blood flow or bleeding in the brain. There are also no problems with the fluid that surrounds the brain.
\vspace{\baselineskip}

Overall, it looks like the mass on the right side of the back of the head has grown quickly and is causing some swelling. We think it might be a type of cancer called glioblastoma. There is no new mass on the right side of the front of the head.
\end{tcolorbox}

\clearpage

\subsubsection{Simplified Report Sample \#13}
\label{app:rep2.13}

\begin{tcolorbox}[]
This medical report is talking about a person's brain and some \hlc{changes that have happened recently.} There was a mass or growth on the right side of the brain that has gotten bigger in a short amount of time. There is also \hlc{some} extra fluid around this mass. On the other side of the brain, there doesn't seem to be any changes. \hlc{There are no signs of problems with the blood or fluid in the brain.} The doctors think that this mass might be a type of brain cancer called glioblastoma.
\end{tcolorbox}

\clearpage

\subsubsection{Simplified Report Sample \#14}
\label{app:rep2.14}

\begin{tcolorbox}[]
This medical report is about a person who has had some medical exams done in the past and has a new exam done recently. In the new exam, there is a mass (a lump or growth) in the back of their head that has gotten bigger since the last exam. The area around the mass is also swollen. There was no evidence of a mass in the front of the head where the person had surgery before. The report also mentions that there is no evidence of bleeding in the brain or problems with the fluid that surrounds the brain. The doctors think that the mass in the back of the head is probably a type of brain cancer called a glioblastoma.
\end{tcolorbox}

\clearpage

\subsubsection{Simplified Report Sample \#15}
\label{app:rep2.15}

\begin{tcolorbox}[]
This medical report is talking about a person's brain and some changes that have happened since their last medical examination. It mentions that there is a mass or growth on the right side of the brain that has gotten bigger since the last examination. There is also some swelling around this mass. The report also mentions that there is no evidence of any other masses or growths in the brain or any problems with the blood or fluid in the brain. The conclusion of the report says that the mass and swelling on the right side of the brain are likely caused by a type of cancer called glioblastoma.
\end{tcolorbox}

\clearpage

\subsection{Report \#3: Oncology CT}
\addtocounter{subsubsection}{-1}

\subsubsection{Original Report}
\label{sec:appendix_original_report3}

\begin{tcolorbox}[]
Findings:\\
Multiple CT scans available for comparison.
\vspace{\baselineskip}

Thorax:\\
Port system right prepectoral. Heart normal in size. No pericardial effusion. No metastatic thoracic lymph nodes. Thyroid struma. No pleural effusions. No pneumonia. No pneumothorax. Marked pleural calcifications unchanged. Pulmonary metastases show response to therapy, e.g., size regrowth of left apical metastasis at 1.9 x 1.5 cm (previously 2.2 x 2.1 cm) and decreasing solid portions with new cavitation in left subpleural metastasis in lower lobe at 1.8 cm (previously 1.9 cm). No new pulmonary metastases detected.
\vspace{\baselineskip}

Abdomen:\\
Homogeneous contrast of liver parenchyma with no metastatic lesions. Portal vein and hepatic veins open. Gallbladder non-irritant with small calculus. No cholestasis. Spleen, pancreas and adrenal glands unremarkable. Lateral contrast of the kidneys. No urinary retention. No ascites. No free abdominal air. Z. n. resection of lymph node metastases in the mesentery, no evidence of recurrence there. No newly metastatic abdominal lymph nodes. Unchanged prominent retroperitoneal lymph nodes, these in need of further control.
\vspace{\baselineskip}

Skeleton:\\
No suspicious osseous lesions. No evidence of acute fractures.
\vspace{\baselineskip}

Conclusion:\\
In metastatic colorectal carcinoma, significant response to therapy in comparison to the previous examination:
\begin{enumerate}
    \item pulmonary metastases present partly in regression and partly constant, but in the latter case with decreased solid parts in the sense of a therapy response.
    \item no evidence of recurrence or new lymph node metastases after resection of the mesenteric lymph node metastasis.
    \item no evidence of new hematogenous distant metastases in the examination area.
\end{enumerate}

\end{tcolorbox}

\clearpage

\subsubsection{Simplified Report Sample \#1}
\label{app:rep3.1}

\begin{tcolorbox}[]
This medical report is talking about a person who has a type of cancer called colorectal carcinoma, which means cancer that starts in the colon or rectum. The report has information about different parts of the body and how the cancer is affecting them.
\vspace{\baselineskip}

The thorax is the area in the body that includes the chest and the organs inside it, like the heart and lungs. The report says that the heart is normal in size and there is no extra fluid around it. It also says that there are no infections in the lungs and no air leaking from the lungs. The report says that there are some spots on the lungs where the cancer has spread, but they are getting smaller and changing in a way that shows that the person is responding to treatment.
\vspace{\baselineskip}

The abdomen is the area in the body that includes the belly and the organs inside it, like the liver, stomach, and intestines. The report says that the liver looks normal and doesn't have any cancer spreading to it. It also says that the gallbladder and spleen look normal, and there is no extra fluid in the belly. The report says that there was surgery to remove some cancer from the lymph nodes in the mesentery, which is a sheet of tissue that holds the intestines in place. The report says that there is no cancer in the lymph nodes now, but there are some other lymph nodes in the belly that need to be checked again.
\vspace{\baselineskip}

The skeleton is the body's framework of bones. The report says that there are no bones with cancer and no broken bones.
\vspace{\baselineskip}

The conclusion of the report is that the person's cancer is responding to treatment. It is getting smaller in some places and changing in a way that shows the treatment is working. There is no new cancer in the lymph nodes or other parts of the body, and the surgery to remove cancer from the lymph nodes was successful.
\end{tcolorbox}

\clearpage

\subsubsection{Simplified Report Sample \#2}
\label{app:rep3.2}

\begin{tcolorbox}[]
This medical report is about a person who has cancer that has spread from their colon to other parts of their body. The report describes the results of some tests that were done to see how the cancer is doing.
\vspace{\baselineskip}

The tests show that there are some cancer cells in the person's lungs, but they are getting smaller. There is also no evidence that the cancer has spread to other parts of the body, like the liver or bones. There were also some cancer cells removed from the person's belly, but they are not growing back.
\vspace{\baselineskip}

Overall, it looks like the person is responding well to treatment, which means that the cancer is getting smaller or not spreading as much. It is important to continue to monitor the cancer and see how it responds to treatment.
\end{tcolorbox}

\clearpage

\subsubsection{Simplified Report Sample \#3}
\label{app:rep3.3}

\begin{tcolorbox}[]
This medical report is about a person who has cancer that has spread to other parts of their body. The report is talking about different parts of the body and whether there are any signs of cancer in those areas.
\vspace{\baselineskip}

In the thorax, or chest, the report says that the person's heart is normal and there is no extra fluid around it. There are no signs of cancer in the lymph nodes in the chest \hlc{or in the thyroid gland.} There is no infection in the lungs and \hlc{no holes in the walls between the lungs and chest.} The report also mentions some spots on the person's lungs that are caused by the cancer. These spots are getting smaller, \hlc{which is a good sign and means that the person is responding to treatment.}
\vspace{\baselineskip}

In the abdomen, or belly, the report says that the liver looks normal and there are no signs of cancer. The gallbladder and bile ducts are also normal. The spleen, pancreas, and adrenal glands are all normal. The kidneys are also normal and there is no extra fluid in the belly. The report mentions some cancer in the lymph nodes in the belly that has been removed, but there are no signs that the cancer has come back or spread to other lymph nodes.
\vspace{\baselineskip}

In the skeleton, or bones, the report says that there are no signs of cancer or any recent fractures.
\vspace{\baselineskip}

Overall, the report says that the person's cancer is responding to treatment. The cancer in the lungs is getting smaller and there are no new signs of cancer in the other parts of the body that were checked.
\end{tcolorbox}

\clearpage

\subsubsection{Simplified Report Sample \#4}
\label{app:rep3.4}

\begin{tcolorbox}[]
A medical report is a document that explains the results of different tests that were done on a person. The report has information about the person's chest (called the thorax), stomach (called the abdomen), and bones (called the skeleton).
\vspace{\baselineskip}

The chest looks normal, and the heart is the right size. There is no extra fluid around the heart or in the lungs, and the person does not have an infection in the lungs. The thyroid gland is swollen, but there are no other problems in the chest. There are some spots on the lungs that are smaller than they were before and are getting better with treatment. There are no new spots on the lungs.
\vspace{\baselineskip}

The stomach looks good, with no abnormal growths. The liver and gallbladder are normal, and there is no blockage in the bile ducts. The spleen, pancreas, and adrenal glands are normal. The kidneys are working well and there is no extra fluid in the stomach. There are no new abnormal growths in the lymph nodes in the stomach, and the lymph nodes in the back of the stomach will need to be checked again later.
\vspace{\baselineskip}

The bones are normal and there are no broken bones.
\vspace{\baselineskip}

Overall, the person's cancer is responding well to treatment. The abnormal growths in the lungs are getting smaller and there are no new abnormal growths in the lymph nodes or other parts of the body.
\end{tcolorbox}

\clearpage

\subsubsection{Simplified Report Sample \#5}
\label{app:rep3.5}

\begin{tcolorbox}[]
A CT scan is a special kind of X-ray that takes pictures of the inside of your body. The doctor looked at multiple CT scans from different times to see how your body has changed.
\vspace{\baselineskip}

The doctor saw that your heart is normal in size and there is no extra fluid around it. The doctor also saw that your thyroid gland is bigger than normal. The thyroid is a small gland in your neck that helps control your body's functions. The doctor did not see any infection in your lungs or any air in the space between your lungs and chest wall. There were some \hlc{spots} on your lungs that have gotten smaller because of the medicine you are taking. There were also some \hlc{spots} on your lungs that have changed, but they are not getting bigger. The doctor did not see any new \hlc{spots} on your lungs.
\vspace{\baselineskip}

The doctor saw that your liver looks normal and there are no new spots on it. The veins that carry blood to and from your liver are open and working properly. Your gallbladder and the tube that carries bile (a substance that helps you digest food) from your liver to your intestine are normal. The spleen, pancreas, and adrenal glands (organs that help with digestion and other body functions) are normal. The doctor saw that your kidneys (organs that help filter waste from your blood) are normal and there is no urine (pee) building up in your bladder. There is no extra fluid in your belly and no extra air in your belly. The doctor saw that the lymph nodes (small organs that help fight infection) near your intestine were removed before and there are no new ones. There are some lymph nodes near your belly that the doctor will keep an eye on.
\vspace{\baselineskip}

The doctor did not see any new or suspicious \hlc{spots} on your bones and there are no broken bones.
\vspace{\baselineskip}

Overall, the doctor saw that the cancer in your body is responding to the medicine you are taking. The cancer in your lungs is partly getting smaller and partly staying the same size, but the parts that are staying the same size are changing. There is no cancer in the lymph nodes near your intestine and no new cancer in other parts of your body.
\end{tcolorbox}

\clearpage

\subsubsection{Simplified Report Sample \#6}
\label{app:rep3.6}

\begin{tcolorbox}[]
This medical report is about a person who has a type of cancer called colorectal carcinoma, which means the cancer started in the colon or rectum. The doctor used some special machines called CT scans to take pictures of the inside of the person's body to see how the cancer is doing.
\vspace{\baselineskip}

The pictures showed that the person's heart is normal and there is no extra fluid around it. The thyroid gland (a small butterfly-shaped gland in the neck) looks a little bigger than usual, but there are no other problems in the chest. There are some spots on the person's lungs where the cancer has spread, but they are getting smaller and changing in a way that shows the person is responding to treatment. There are no new spots on the lungs.
\vspace{\baselineskip}

The pictures also showed that the person's liver looks normal and there are no cancer spots on it. The gallbladder (a small organ that stores bile) has a small stone, but it's not causing any problems. The rest of the organs in the stomach area look normal too. There was some cancer in the lymph nodes (small, bean-shaped organs that help fight infections) near the stomach, but it has been removed and there is no sign that it has come back. There are some bigger lymph nodes in the \hlc{back of the stomach} that the doctor will keep an eye on.
\vspace{\baselineskip}

The pictures didn't show any problems with the person's bones, and there are no broken bones. Overall, the doctor thinks that the person's cancer is responding well to treatment and there is no new cancer in the body.
\end{tcolorbox}

\clearpage

\subsubsection{Simplified Report Sample \#7}
\label{app:rep3.7}

\begin{tcolorbox}[]
Here is a summary of the medical report in simple language:
\vspace{\baselineskip}

The doctor did some special pictures of the inside of the person's body using a machine called a \hlc{CT scan.} They looked at the person's chest and saw that the heart was a normal size and there was no fluid around it. They also saw that the person had an \hlc{extra thyroid gland,} but there was no infection in the chest and no air was leaking out of the lungs. The doctor saw that the cancer in the person's lungs was getting smaller after treatment and there were no new areas of cancer.
\vspace{\baselineskip}

The doctor also looked at the \hlc{person's stomach} and saw that the liver was healthy and there were no cancer cells there. The gallbladder was also healthy, with just a small stone inside. The other organs in the stomach, like the spleen and pancreas, were normal. The kidneys looked good and there was no extra fluid in the stomach. The doctor saw that the cancer in the person's abdomen had been removed and there was no evidence that it had come back. There were also no new areas of cancer in the abdomen.
\vspace{\baselineskip}

The doctor also looked at the person's bones and saw that they were normal and not broken.
\vspace{\baselineskip}

Overall, the doctor saw that the treatment for the person's cancer was working and the cancer was getting smaller. There were no new areas of cancer and no evidence that the cancer had come back.
\end{tcolorbox}

\clearpage

\subsubsection{Simplified Report Sample \#8}
\label{app:rep3.8}

\begin{tcolorbox}[]
The medical report is about a person who has cancer that has spread to other parts of their body. The doctors did some special pictures of the inside of the person's body \hlc{to see how the cancer is doing.} The cancer in the person's lungs is getting smaller in some parts and there is less solid cancer in other parts, which is good because it means the medicine the person is taking is working. The cancer has not spread to any new parts of the body and the cancer that was removed from the person's intestine is not coming back. The bones are all healthy and there are no broken bones.
\end{tcolorbox}

\clearpage

\subsubsection{Simplified Report Sample \#9}
\label{app:rep3.9}

\begin{tcolorbox}[]
This medical report is talking about a person who has cancer in their colon (part of the large intestine) that has spread to other parts of their body. The doctors used a special machine called a CT scan to take pictures of the inside of the person's body to see how the cancer is doing. The CT scan showed that the cancer in the person's lungs is getting smaller in some parts and not growing as much in other parts, \hlc{which is a good sign that the treatment is working}. The CT scan also showed that there is no cancer in the person's liver, gallbladder, spleen, pancreas, adrenal glands, or kidneys, and that there is no cancer in the person's bones. There is also no cancer in the person's abdomen (belly) after the doctors removed some lymph nodes that had cancer in them. However, there are some lymph nodes \hlc{in the person's back that the doctors need to} keep an eye on because they might have cancer. Overall, the treatment seems to be helping the person's cancer get better.
\end{tcolorbox}

\clearpage

\subsubsection{Simplified Report Sample \#10}
\label{app:rep3.10}

\begin{tcolorbox}[]
This medical report is about a person who has cancer that has spread from the colon to other parts of the body. The report is talking about a series of tests that were done to see how the cancer is doing.
\vspace{\baselineskip}

The tests show that the person's heart is a normal size and is not causing any problems. The thyroid gland is swollen, but there is no fluid or infection around the lungs. The lungs have some cancer in them, but the cancer is getting smaller and there is no new cancer. The liver and other organs in the belly are normal and there are no new cancer cells in the belly. The bones do not have any cancer and there are no new broken bones.
\vspace{\baselineskip}

Overall, the cancer is responding to treatment and is not getting worse or spreading to new places. There are still some cancer cells in the lungs, but they are getting smaller. The cancer surgery in the belly was successful and there is no evidence of the cancer coming back in that area. There are no new cancer cells in any other parts of the body.
\end{tcolorbox}

\clearpage

\subsubsection{Simplified Report Sample \#11}
\label{app:rep3.11}

\begin{tcolorbox}[]
This medical report is about a person who has a type of cancer called colorectal carcinoma. The report mentions some scans that have been taken of the person's body.
\vspace{\baselineskip}

The scans show that the person's heart is normal and there are no problems with the fluid around the heart. The thyroid gland looks swollen, but there are no signs of infection in the lungs or fluid in the space between the lungs and the chest wall. The cancer in the lungs looks like it is getting smaller and there are no new areas of cancer in the lungs.
\vspace{\baselineskip}

The scans also show that the liver looks normal and there are no new areas of cancer in the abdomen. The gallbladder has a small stone, but there is no blockage of bile. The spleen, pancreas, and adrenal glands all look normal. The kidneys are normal and there is no urine build-up in the bladder. There is no fluid in the abdomen and there is no air in the abdomen. There was surgery to remove cancer in the lymph nodes in the abdomen and there is no evidence that the cancer has come back there. There are some swollen lymph nodes in the back of the abdomen that need to be watched closely.
\vspace{\baselineskip}

The scans also show that there are no new or unusual areas of cancer in the bones. There are no broken bones.
\vspace{\baselineskip}

Overall, the cancer in the lungs looks like it is getting smaller and there is no evidence that the cancer has spread to other parts of the body or come back after surgery.
\end{tcolorbox}

\clearpage

\subsubsection{Simplified Report Sample \#12}
\label{app:rep3.12}

\begin{tcolorbox}[]
This medical report is about a person who has cancer that has spread to different parts of their body. The doctors used special pictures (CT scans) to look at the person's body and check for any changes. They looked at the person's chest and saw that their heart is normal and there is no fluid around it. They also saw that the person has \hlc{thyroid problems}, but no fluid in their lungs or any signs of pneumonia. The cancer in the person's lungs has \hlc{gotten smaller and is not spreading as much as before}. The doctors also looked at the person's abdomen (stomach) and saw that their liver, gallbladder, spleen, pancreas, and kidneys are all normal. They also saw that the cancer in the person's stomach has not spread to other parts of the body and has not come back after surgery. Finally, the doctors looked at the person's bones and saw that there are no new cancer spots or broken bones. Overall, it looks like the person's cancer is responding well to treatment and is not getting worse.
\end{tcolorbox}

\clearpage

\subsubsection{Simplified Report Sample \#13}
\label{app:rep3.13}

\begin{tcolorbox}[]
This medical report is talking about the results of some tests that were done on a person's body. The tests included taking pictures of the inside of their body using a special machine called a CT scan. The report mentions different parts of the body, such as the thorax (the chest), the abdomen (the belly), and the skeleton (the bones).
\vspace{\baselineskip}

The report says that the heart is normal in size and there is no fluid around it. It also says that the person has an \hlc{infection} in their thyroid gland and there is no fluid in their chest or belly. The report mentions some spots on the person's lungs that are caused by cancer. These spots have gotten smaller since the last time they were checked, which means that the treatment the person is getting is working. There are no new cancer spots on the person's lungs.
\vspace{\baselineskip}

The report also mentions the person's liver, which is a big organ in the belly that helps with digestion and getting rid of waste. It says that the liver looks normal and there are no cancer spots on it. The report also mentions the person's gallbladder, spleen, pancreas, and kidneys, which are all organs in the belly that help with different functions in the body. It says that these organs look normal. The report also mentions some lymph nodes, which are small bumps that help fight infections. It says that some lymph nodes in the person's belly have been removed because they had cancer, but there is no evidence that the cancer has come back. The report also mentions the person's bones and says that there are no unusual spots on them and no broken bones.
\vspace{\baselineskip}

Overall, the report says that the person has cancer that started in their colon (a part of the digestive system) and has spread to other parts of their body. The report says that the treatment the person is getting is working and there \hlc{is no evidence of the cancer spreading} to other parts of the body.
\end{tcolorbox}

\clearpage

\subsubsection{Simplified Report Sample \#14}
\label{app:rep3.14}

\begin{tcolorbox}[]
This medical report is talking about some pictures that were taken of the inside of a person's body. The pictures show how the person's body is doing and how it has changed since the last time they got pictures taken.
\vspace{\baselineskip}

The pictures show that the person's heart is normal and there is no extra fluid around it. The person's thyroid gland is bigger than usual, but there is no extra fluid in their chest and they do not have an infection in their lungs. The person has some hard lumps on the inside of their chest that have not changed. They also have some spots on their lungs that were caused by cancer. These spots have gotten smaller and have some holes in them, which is a good sign that the person is getting better.
\vspace{\baselineskip}

The pictures also show that the person's liver looks normal and there are no cancer spots on it. The person's gallbladder has a small stone, but it is not causing any problems. The person's other organs, like the spleen and pancreas, look normal. The person's kidneys and urinary system are also normal and there is no extra fluid in the person's belly. The person had surgery to remove some cancer spots in their intestine and there are no new cancer spots there. However, there are some hard lumps on the person's back that doctors will keep an eye on.
\vspace{\baselineskip}

Finally, the pictures do not show any new bones with cancer or any broken bones. Overall, the person's body is showing a good response to treatment and the cancer is getting better in some areas.
\end{tcolorbox}

\clearpage

\subsubsection{Simplified Report Sample \#15}
\label{app:rep3.15}

\begin{tcolorbox}[]
This medical report is about a person who has cancer that has spread from their colon to other parts of their body. The doctors used special pictures called CT scans to see inside the person's body. They looked at the person's chest and saw that their heart is normal and not causing any problems. They also saw some changes in the cancer in the person's lungs that suggest the treatment the person is receiving is working. The doctors also looked at the person's abdomen and saw that the cancer has not spread to the liver or other organs there. They saw some changes in the lymph nodes in the person's belly that may need to be watched closely. Finally, the doctors looked at the person's bones and did not see any signs of cancer or broken bones. Overall, it looks like the treatment the person is receiving is helping to slow down the cancer.
\end{tcolorbox}

\clearpage

\section{Questionnaire}
\label{app:questionnaire}
\subsection{Questionnaire Design}
\label{app:questionnaire_questions}
\begin{figure}[h]
\begin{tcolorbox}[standard jigsaw, opacityback=0,]
\centering
\includegraphics[width=\textwidth, trim={1.5cm 9cm 1.5cm 2.5cm},clip]{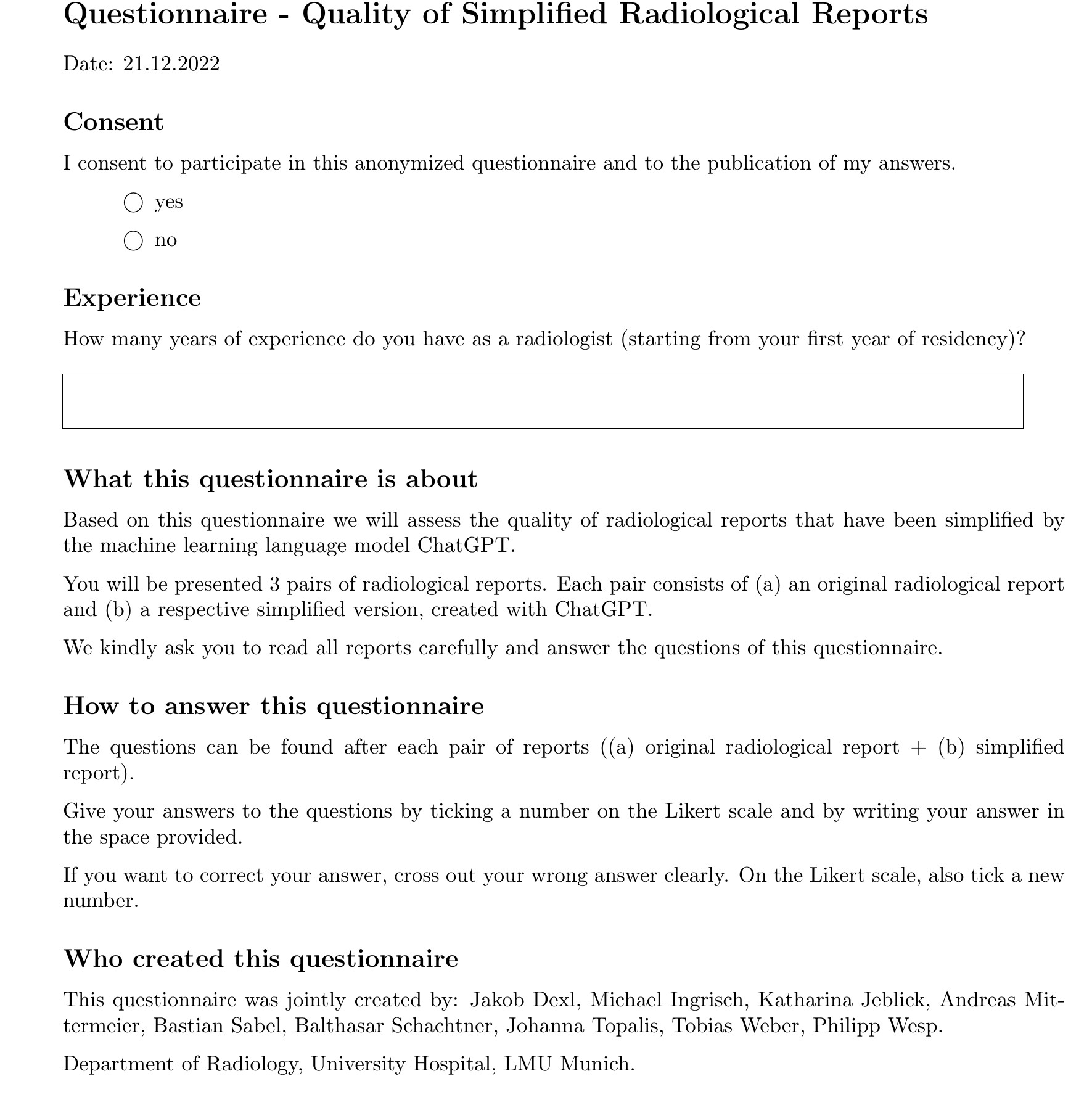}
\end{tcolorbox}
\caption{Meta questions and cover for the questionnaire.}
\end{figure}
\clearpage
\begin{figure}[h]
\begin{tcolorbox}[standard jigsaw, opacityback=0,]
\centering
\includegraphics[width=\textwidth, trim={1.5cm 5cm 1.5cm 2.5cm},clip]{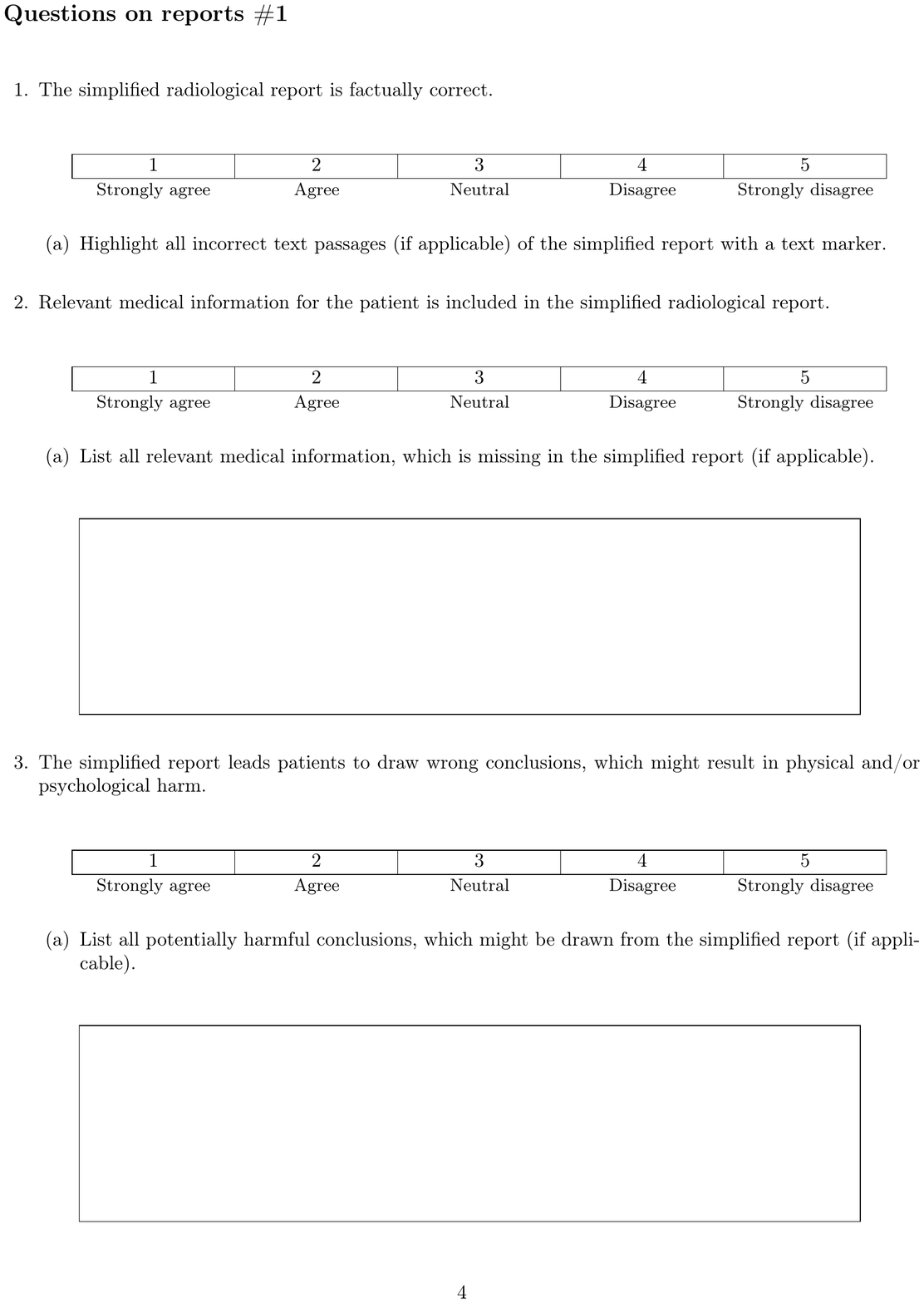}
\end{tcolorbox}
\caption{Questions concerning one of the simplified reports provided to the radiologist in the questionnaire.}
\end{figure}

\subsection{Answers to the Questionnaire}
\label{app:questionnaire_answers}
\begin{landscape}
\begin{longtable}[c]{@{}lllllp{3.6cm}lp{3.6cm}lp{3.6cm}@{}}
\toprule
Participant & Experience & Sample & Report & Q1 & Q1a & Q2 & Q2a & Q3 & Q3a \\* \midrule
\endhead
\bottomrule
\endfoot
\endlastfoot
\rowcolor[HTML]{EFEFEF} 
1 & 4.00 & \ref{app:rep1.1} & 1 & 1 & none & 1 & none & 4 & none \\
\rowcolor[HTML]{EFEFEF} 
 &  & \ref{app:rep2.1} & 2 & 3 & There is no new mass in the right side of the brain & 2 & none & 2 & contradicting results (new brain lesion vs. no new brain lesion) \\
\rowcolor[HTML]{EFEFEF} 
 &  & \ref{app:rep3.1} & 3 & 1 & none & 1 & none & 5 & none \\
2 & 1.00 & \ref{app:rep1.2} & 1 & 2 & none & 1 & none & 4 & none \\
 &  & \ref{app:rep2.2} & 2 & 2 & none & 2 & no intermediate or recent ischemia & 4 & glioblastoma in conclusion, too specific; it's a DD not yet diagnosed \\
 &  & \ref{app:rep3.2} & 3 & 2 & none & 1 & none & 4 & none \\
\rowcolor[HTML]{EFEFEF} 
3 & 0.50 & \ref{app:rep1.3} & 1 & 2 & none & 2 & none & 4 & none \\
\rowcolor[HTML]{EFEFEF} 
 &  & \ref{app:rep2.3} & 2 & 2 & small; bigger & 3 & none & 3 & significant growth (almost doubled) should maybe not be played down with the word "small" growth and I would also use the word significant in the conclusion \\
\rowcolor[HTML]{EFEFEF} 
 &  & \ref{app:rep3.3} & 3 & 2 & or in the thyroid gland; no holes in the walls between the lungs and chest; which is a good sign and means that the person is responding to treatment & 2 & none & 3 & I would leave out the conclusion that it "is a good sign and means that a person is responding to the treatment" or move it to the last paragraph of the report \\
4 & 3.50 & \ref{app:rep1.4} & 1 & 1 & and tear; the area between the two bones that make up the lower part of & 2 & none & 4 & none \\
 &  & \ref{app:rep2.4} & 2 & 2 & swelling & 4 & in conclusion growth of the lesion parietooccipital is missing & 3 & information in conclusion about the lesion showing growth is missing \\
 &  & \ref{app:rep3.4} & 3 & 2 & none & 2 & none & 4 & none \\
\rowcolor[HTML]{EFEFEF} 
5 & 6.00 & \ref{app:rep1.5} & 1 & 2 & In the middle part of your leg; middle part of the leg & 1 & cartilage damage is on the inner side of the knee (thigh bone) & 5 & none \\
\rowcolor[HTML]{EFEFEF} 
 &  & \ref{app:rep2.5} & 2 & 1 & none & 1 & none & 3 & none \\
\rowcolor[HTML]{EFEFEF} 
 &  & \ref{app:rep3.5} & 3 & 4 & spots; spots; spots; spots & 2 & spot vs metastasis; decreasing solid portions in pulmonal metastasis & 2 & "the parts that are staying the same size are changing" unclear if good or bad; not clear that spots are pulmonal metastasis \\
6 & 5.00 & \ref{app:rep1.6} & 1 & 2 & none & 2 & infraction lateral tibia plateau & 4 & none \\
 &  & \ref{app:rep2.6} & 2 & 2 & abnormal; dye; is & 2 & none & 3 & DD $\neq$ "is" \\
 &  & \ref{app:rep3.6} & 3 & 2 & back of the stomach & 2 & none & 4 & none \\
\rowcolor[HTML]{EFEFEF} 
7 & 10.00 & \ref{app:rep1.7} & 1 & 2 & in the bone inside your lower leg bone; to the muscle in the back of your knee; top of your thigh bone & 2 & none & 4 & none \\
\rowcolor[HTML]{EFEFEF} 
 &  & \ref{app:rep2.7} & 2 & 2 & head; head & 2 & second/ distant manifestation of known GBM in right frontal lobe which has been resected & 3 & GBM is one (likely) DD which implies that other DDs exist (e.g. radionecrosis). This uncertainty is not conveyed in the simplyfied report \\
\rowcolor[HTML]{EFEFEF} 
 &  & \ref{app:rep3.7} & 3 & 2 & CT scan; extra thyroid gland; person's stomach & 1 & none & 4 & none \\
8 & 0.40 & \ref{app:rep1.8} & 1 & 3 & none & 2 & none & 3 & none \\
 &  & \ref{app:rep2.8} & 2 & 4 & think that this mass might be a type of brain tumor called a glioblastoma (GBM) & 2 & none & 3 & lacking communication that it might be another (not GBM) brain tumor and weighing of possibilities \\
 &  & \ref{app:rep3.8} & 3 & 4 & to see how the cancer is doing & 4 & none & 2 & lacking communication about the limitations of CT and the factual relevance of size decrease of metastases \\
\rowcolor[HTML]{EFEFEF} 
9 & 4.00 & \ref{app:rep1.9} & 1 & 2 & inside; outside; hard, smooth & 1 & none & 4 & none \\
\rowcolor[HTML]{EFEFEF} 
 &  & \ref{app:rep2.9} & 2 & 2 & does not seem to be damaged or & 2 & none & 2 & brain not damaged! \\
\rowcolor[HTML]{EFEFEF} 
 &  & \ref{app:rep3.9} & 3 & 4 & which is a good sign that the treatment is working; in the person's back that the doctors need to & 2 & none & 2 & lymph nodes \\
10 & 5.00 & \ref{app:rep1.10} & 1 & 2 & none & 1 & none & 4 & none \\
 &  & \ref{app:rep2.10} & 2 & 2 & head; head; front of the head & 2 & none & 4 & none \\
 &  & \ref{app:rep3.10} & 3 & 2 & none & 2 & none & 4 & none \\
\rowcolor[HTML]{EFEFEF} 
11 & 11.00 & \ref{app:rep1.11} & 1 & 2 & none & 2 & none & 3 & none \\
\rowcolor[HTML]{EFEFEF} 
 &  & \ref{app:rep2.11} & 2 & 2 & none & 2 & localisation of the present tumor versus localisation of the excised tumor & 2 & misunderstanding which lesion is stable and which one is in progress can lead the patient to some wrong expectations \\
\rowcolor[HTML]{EFEFEF} 
 &  & \ref{app:rep3.11} & 3 & 1 & none & 1 & none & 4 & none \\
12 & 0.50 & \ref{app:rep1.12} & 1 & 2 & none & 2 & none & 4 & none \\
 &  & \ref{app:rep2.12} & 2 & 2 & none & 2 & none & 3 & none \\
 &  & \ref{app:rep3.12} & 3 & 3 & thyroid problems; gotten smaller and is not spreading as much as before & 2 & none & 3 & none \\
\rowcolor[HTML]{EFEFEF} 
13 & 22.00 & \ref{app:rep1.13} & 1 & 2 & on the outside of the tibia & 2 & none & 4 & none \\
\rowcolor[HTML]{EFEFEF} 
 &  & \ref{app:rep2.13} & 2 & 4 & changes that have happened recently; some; There are no signs of problems with the blood or fluid in the brain & 3 & none & 2 & changes happened recently (how recently?); some extra fluid $\neq$ sign. incr. edema \\
\rowcolor[HTML]{EFEFEF} 
 &  & \ref{app:rep3.13} & 3 & 4 & infection; no evidence of the cancer spreading & 2 & not thyroid infection but enlargement; there is spread of cancer -> lungs & 2 & there is no (new) spread of cancer \\
14 & 8.00 & \ref{app:rep1.14} & 1 & 1 & none & 1 & none & 5 & none \\
 &  & \ref{app:rep2.14} & 2 & 1 & none & 1 & none & 5 & none \\
 &  & \ref{app:rep3.14} & 3 & 3 & none & 2 & back is not specific enough or wrong as the retroperitoneum is actually affected & 2 & patient might be misled to look at cancer spots on their back \\
\rowcolor[HTML]{EFEFEF} 
15 & 10.00 & \ref{app:rep1.15} & 1 & 2 & medial compartment; your body; your leg bone; leg bone; lateral compartment is the part of the body that's on the outside & 2 & report is about the knee & 4 & none \\
\rowcolor[HTML]{EFEFEF} 
 &  & \ref{app:rep2.15} & 2 & 1 & none & 1 & none & 5 & none \\
\rowcolor[HTML]{EFEFEF} 
 &  & \ref{app:rep3.15} & 3 & 1 & none & 1 & none & 5 & none \\* \bottomrule
\caption{Answers of the radiologists in the questionnaires.}
\label{app:tab:questionnaire_answers} \\
\end{longtable}
\end{landscape}
\end{document}